\documentclass[preprint,11pt]{elsarticle}

\usepackage[top=1in, bottom=1in, left=1in, right=1in]{geometry}
\usepackage{amssymb}
\usepackage{amsmath}
\usepackage{hyperref}
\usepackage{amsthm}
\usepackage{graphicx} 
\usepackage{xcolor}
\usepackage{multirow}
\usepackage{comment}
\usepackage{caption}
\usepackage{subcaption}
\usepackage{relsize}
\usepackage{algorithm}
\usepackage{algpseudocode}
\usepackage{subcaption}
\usepackage{fancyhdr}
\usepackage{mathtools}
\newtheorem{theorem}{Theorem}
\theoremstyle{plain}
\newtheorem{remark}{Remark}
\journal{Journal of Computational Physics}

\begin{document}
\begin{frontmatter}
\title{Federated Learning with Uncertainty Quantification on Pseudo Label Exchange}
\author[label1]{Jingqiao Tang} 
\author[label1]{Phuoc-Toan Huynh}
\author[label2]{Richard Archibald \corref{cor1}}
\author[label1]{Feng Bao}

\affiliation[label1]{organization={Department of Mathematics, Florida State University},
            addressline={222 S Copeland St}, 
            city={Tallahassee},
            postcode={32306}, 
            state={FL},
            country={USA}}

\affiliation[label2]{organization={Oak Ridge National Laboratory},
            addressline={1 Bethel Valley Road}, 
            city={Oak Ridge},
            postcode={37830}, 
            state={TN},
            country={USA}}

\cortext[cor1]{Corresponding author}

\begin{abstract}
In this work, we investigate federated stochastic operator learning for stochastic partial differential equations (SPDEs), with an emphasis on heterogeneous client data distributions and uncertainty quantification. Federated learning enables clients to collaboratively train models without exchanging their original solution data, but conventional federated averaging can suffer significant performance degradation when client distributions differ, particularly when their input function supports are disjoint. To address these challenges, we propose the Federated Stochastic Operator Network (FedSON), a two-phase framework that combines auxiliary data construction with stochastic operator learning. In the first phase, locally trained deterministic neural operators and their corresponding input functions are exchanged to generate pseudolabeled auxiliary data, enabling the federated model to learn across heterogeneous input distributions without sharing original solution trajectories. In the second phase, a stochastic operator network is trained on top of the frozen deterministic operator to characterize stochastic fluctuations and quantify uncertainty in the predicted solutions. Numerical experiments involving the heat equation, viscous Burgers' equation, and two-dimensional advection–diffusion equation demonstrate that FedSON substantially improves deterministic prediction accuracy under heterogeneous client distributions while producing uncertainty estimates comparable to those obtained through centralized training.
\end{abstract}

\begin{keyword}

Federated learning,  Operator learning, Stochastic partial differential equations, DeepONet, Stochastic Operator Networks, uncertainty quantification, Stochastic maximum principle

\end{keyword}
\end{frontmatter}

\section{Introduction}\label{sec:intro}

Modern scientific and engineering data are increasingly generated and stored in a distributed fashion, across many institutions, instruments, or devices rather than in a single central repository. There are several reasons why such data are not simply pooled. Privacy and proprietary constraints often prohibit sharing: hospitals, companies, and national laboratories may be unable to release raw measurements even when they are willing to collaborate on a shared model~\citep{kairouz2021advances, sheller2020medicine}. The massive volume of modern datasets can make centralized collection impractical, as transmitting large volumes of data to a central server is costly in bandwidth and storage~\citep{konecny2016strategies}.In addition, the rise of edge computing places computation close to where data originate, so that learning is naturally performed on distributed devices rather than after aggregation~\citep{nguyen2021iot, shi2016edge}. These considerations motivate collaborative learning paradigms that keep data local.

Among the approaches developed for collaborative learning with decentralized data, federated learning (FL) has emerged as a widely adopted framework~\citep{kairouz2021advances,li2020federated, mcmahan2017fedavg}. Rather than moving data to the model, FL moves the model to the data: each client trains on its own local dataset and communicates model updates to a central server, which aggregates them into a shared global model. By keeping raw training data local, FL can reduce the need for centralized data collection and has attracted considerable interest in privacy-sensitive and bandwidth-limited applications, including healthcare, finance, and the Internet of Things~\citep{kairouz2021advances,nguyen2021iot,sheller2020medicine}. One of the most widely used aggregation algorithms is Federated Averaging (FedAvg), which combines locally trained models through a data-size-weighted average and has demonstrated effectiveness in distributed learning~\citep{bonawitz2019towards, mcmahan2017fedavg}.

A central obstacle to this approach is data heterogeneity. When the local data distributions differ across clients, the so-called non-IID setting, the average of locally trained models can depart substantially from the model that centralized training would have produced, and the accuracy of FedAvg can degrade~\citep{karimireddy2020scaffold,li2020convergence,zhao2018noniid}. Heterogeneity is one of the most actively studied challenges in federated learning, and a range of algorithmic approaches has been proposed to mitigate it~\citep{karimireddy2020scaffold,li2020fedprox,wang2020fednova,zhu2021datafree}. In many settings, increasing data heterogeneity further degrades federated learning performance, making the extreme, strongly non-IID case particularly challenging~\citep{hsu2019measuring}.

In this work, we study federated learning in the context of partial differential equations (PDEs), which govern a vast range of physical systems in science and engineering. Rather than learning an end-to-end map for a single problem instance, our goal is to learn the underlying physics in the form of a \emph{solution operator}: a map from a problem input, such as a forcing term or an initial condition, to the corresponding solution function. This is the objective of \emph{operator learning}, which approximates maps between infinite dimensional function spaces and, once trained, predicts the solution for a new input in a single forward pass without re-solving the equation~\citep{Bhattacharya2021, chen1995universal, Guo2024,  kovachki2023neural, li2021fno, lu2021deeponet, Rahman2022}. Operator learning is thus a natural target for the distributed scientific setting, where different clients may hold solution data corresponding to different regions of the input space.

Beyond recovering the solution operator, scientific data are rarely noise-free. Measurement errors, stochastic forcing, and uncertainties in physical models can introduce variability into the observed solution trajectories. In a federated setting, these uncertainties may differ across clients because of variations in their measurement processes and local environments. Consequently, a deterministic neural operator that produces a single prediction may be insufficient for characterizing the variability present in the data. This motivates the development of stochastic operator learning methods that can capture both the underlying solution structure and the uncertainty associated with the predicted solutions.

Several approaches have been developed to incorporate uncertainty quantification (UQ) into operator learning, including Bayesian methods, ensemble-based approaches, and direct predictive uncertainty estimation~\citep{lin2023bdeeponet,Winovich2026,Winovich2019ConvPDEUQ,yang2022randomized}. While Bayesian and ensemble-based methods can provide meaningful uncertainty estimates, posterior sampling and ensemble training may introduce substantial computational overhead. An alternative is the Stochastic Operator Network (SON), which incorporates stochastic dynamics into neural operator learning to generate probabilistic solution predictions~\citep{bausback2025son}. However, jointly optimizing the deterministic operator and stochastic components can be computationally demanding, particularly for complex problems requiring large neural network architectures. To address this challenge, a two-phase training strategy was developed in~\cite{huynh2026diffusion} which separates the learning of the deterministic solution operator from the modeling of stochastic fluctuations. In the first phase, a deterministic neural operator is trained to capture the primary solution structure. In the second phase, the deterministic operator is frozen, and a stochastic network is trained to characterize the variability around its predictions. This separation simplifies the stochastic training stage by avoiding the simultaneous optimization of the full deterministic and stochastic architectures, while retaining the ability to generate probabilistic predictions.

In this paper, we bring these challenges together and propose the \emph{Federated Stochastic Operator Network} (FedSON), a two-phase framework for learning deterministic solution structures and stochastic fluctuations from distributed, heterogeneous data without exchanging original solution trajectories. The proposed framework combines an auxiliary data mechanism for addressing heterogeneous input distributions with a stochastic operator network for uncertainty quantification. Our main contributions are as follows: First, we study federated operator learning in the extreme case in which the clients' input function supports are \emph{fully disjoint}, a setting that, to our knowledge, has not been systematically investigated, and we show that standard federated averaging fails there. Second, we propose an auxiliary data mechanism based on pseudolabel exchange, in which clients share locally trained models and input functions, but never their solution data, to reconstruct the missing regions of the input space and restore the performance of federated averaging. Third, we develop a federated two-phase stochastic operator learning strategy in which the deterministic operator is first trained using the auxiliary data mechanism and subsequently frozen, while a stochastic network learns the variability in the noisy solution trajectories. This approach enables uncertainty quantification without requiring clients to exchange their original solution data. We demonstrate the approach on four PDE problems: the heat equation, the viscous Burgers' equation, and the one and two dimensional advection diffusion equations, and compare throughout against homogeneous and centralized references.

The remainder of the paper is organized as follows: Section~\ref{sec:fl} reviews federated learning and the data heterogeneity problem. Section~\ref{sec:operator-learning} introduces operator learning and its stochastic extension. Section~\ref{sec:fedson} presents the FedSON framework, including the pseudolabel exchange mechanism and the two-phase federated training. Section~\ref{sec:experiments} reports the numerical experiments, and Section~\ref{sec:conclusion} concludes.

\section{Federated Learning}\label{sec:fl}

Federated learning (FL) is a distributed machine learning paradigm in which many clients collaboratively train a shared model under the coordination of a central server, while their local training data remain on their own devices~\citep{mcmahan2017fedavg}. As outlined in Section~\ref{sec:intro}, this design keeps data local and is well suited to the distributed, privacy constrained scientific setting studied in this work. In this section, we review the standard federated averaging algorithm (Section~\ref{sec:fl-fedavg}), the data heterogeneity problem that motivates our approach (Section~\ref{sec:fl-hetero}), and prior work on federated operator learning (Section~\ref{sec:fl-operator})

\subsection{Federated Averaging}\label{sec:fl-fedavg}

The standard FL algorithm is Federated Averaging (FedAvg) developed in~\cite{mcmahan2017fedavg}. Consider $K$ clients, where client $k$ holds a local dataset $\mathcal{D}_k$ of size $N_k$, and let $N = \sum\limits_{k=1}^{K} N_k$. FedAvg seeks the global parameters $\theta$ that minimize the aggregate objective
\begin{equation}\label{eq:fl-objective}
\min_{\theta}\; \mathcal{L}(\theta) \;=\; \sum_{k=1}^{K} \frac{N_k}{N}\,\mathcal{L}_k(\theta), \qquad \mathcal{L}_k(\theta) \;=\; \frac{1}{N_k}\sum_{d \in \mathcal{D}_k}\ell(\theta; d),
\end{equation}
where $\mathcal{L}_k$ is the local loss of client $k$ and $\ell(\theta;d)$ is the per sample loss. Training proceeds in communication rounds. At the start of round $l$, the server broadcasts the current global model $\theta^{l}$ to the clients. Each client initializes a local copy $\theta^{l}_k \leftarrow \theta^{l}$ and performs $E$ epochs of local optimization (typically SGD or Adam) on its own data. The server then aggregates the updated local models by a size-weighted average,
\begin{equation}\label{eq:fl-fedavg}
\theta^{l+1} \;\leftarrow\; \sum_{k=1}^{K} \frac{N_k}{N}\,\theta^{l,E}_k,
\end{equation}
where $\theta^{l,E}_k$ denotes client $k$'s model after $E$ local epochs in round $l$. Since clients exchange only parameters, and each performs several local steps between communications, FedAvg reduces the number of communication rounds required to train a model by orders of magnitude relative to naively synchronizing every gradient step~\citep{konecny2016strategies, mcmahan2017fedavg}. Communication efficiency remains a central concern in FL, and a range of methods have been proposed to reduce it further, including update compression and adaptive server optimizers~\citep{kairouz2021advances, konecny2016strategies}.

Sharing model updates rather than training data mitigates, but does not eliminate, privacy risks, as the shared updates can reveal information about the underlying training data~\cite{zhu2019deep}. Complementary mechanisms such as secure aggregation~\citep{bonawitz2017secure}, which enables the server to compute aggregated updates without accessing individual contributions, and differential privacy~\citep{geyer2017dp}, which provides quantifiable privacy guarantees, can be incorporated into FL to strengthen privacy protection. These privacy-preserving mechanisms are beyond the scope of the present work; we adopt standard FedAvg as the aggregation rule throughout and focus on a different obstacle to collaborative training, namely data heterogeneity.

\subsection{Data Heterogeneity and the Convergence Problem}\label{sec:fl-hetero}


FedAvg is well understood in the IID case, where the weighted average of locally trained models approximates the model that centralized training on the pooled data would produce, and convergence is favorable~\citep{mcmahan2017fedavg}. The situation changes markedly under heterogeneous, non-IID data, whose effect on FedAvg has been the subject of extensive and, notably, not fully settled study.

A large body of work documents that heterogeneity degrades FedAvg. \citet{zhao2018noniid} showed that accuracy can fall dramatically under highly skewed client distributions and attributed the degradation to \emph{weight divergence}: local models initialized identically drift toward distinct local optima, so their average departs from the centralized solution. They quantified the degree of heterogeneity by the $1$-Wasserstein distance, the minimum ''transport cost'' of reshaping one probability distribution into another, between each client’s data distribution and the pooled population distribution, and showed that the accuracy drop grows with this distance. \citet{karimireddy2020scaffold} gave a complementary analysis, proving that FedAvg suffers from \emph{client drift} under heterogeneity and that this drift persists even with full batch gradients and full client participation. A range of algorithmic remedies followed. \citet{zhao2018noniid} proposed sharing a small globally common subset of data to reduce divergence at the cost of relaxing the data locality constraint. FedProx \citep{li2020fedprox} adds a proximal term that keeps local updates close to the global model, improving stability in highly heterogeneous settings, while SCAFFOLD \citep{karimireddy2020scaffold} uses control variates to correct client drift directly. {A distinct line of work transfers knowledge between clients through model outputs rather than model parameters. Federated knowledge distillation methods, including data-free
variants that require no shared proxy data~\citep{zhu2021datafree}, leverage the knowledge of local models to improve global model training and can outperform parameter averaging under strong
data heterogeneity. In this work, we develop a pseudolabel exchange strategy within this knowledge transfer framework,
tailored to the operator learning setting, as described in Section~\ref{sec:fedson}.}


{Nevertheless, data heterogeneity does not necessarily prevent FedAvg from converging. For example, \citet{li2020convergence} established convergence guarantees for FedAvg under certain assumptions on non-IID data. However, the performance of FedAvg can still be affected by the degree and structure of data heterogeneity across clients. To investigate this effect, we consider four training settings in our numerical experiments: \emph{\textbf{heterogeneous}}, \emph{\textbf{homogeneous}}, \emph{\textbf{auxiliary}}, and \emph{\textbf{centralized}}. The heterogeneous and homogeneous settings allow us to examine the effect of different client data distributions under FedAvg, while the auxiliary setting evaluates the proposed pseudolabel exchange strategy. The centralized setting serves as a reference in which training is performed directly on the pooled data without federated aggregation. These comparisons are presented in Section~\ref{sec:experiments}.

}

\subsection{Federated Learning for Operator Learning}\label{sec:fl-operator}

The integration of federated learning with scientific machine learning is recent and comparatively underexplored. \citet{moya2022feddeeponet} introduced Fed-DeepONet, the first federated training framework for deep operator networks, and observed that naive federated averaging can converge poorly for operator learning tasks. \citet{zhang2024fedsciml} conducted a systematic study of federated scientific machine learning, proposing federated formulations of both physics-informed neural networks and DeepONets, and using the $1$-Wasserstein distance to quantify heterogeneity; for operator learning, they control the non-IID level by sampling input functions from Chebyshev polynomial spaces with differing numbers of active modes, and report that the federated model recovers centralized performance as heterogeneity decreases. A related but architecturally distinct approach, D2NO \citep{zhang2024d2no}, handles heterogeneous input function spaces by assigning a dedicated branch network to each client's subspace and sharing a single consensus trunk network, rather than averaging a common global model.

In contrast to these studies, which retain some overlap between client input distributions, we study the extreme case in which the supports are fully disjoint. The construction of this partition and the experiments that isolate the resulting failure of federated averaging, together with the auxiliary data remedy, are presented in Section~\ref{sec:experiments}.

\section{Operator Learning}\label{sec:operator-learning}

\subsection{Introduction to Operator Learning}\label{sec:ol-intro}

As introduced in Section~\ref{sec:intro}, operator learning aims to approximate \emph{operators}, maps between infinite dimensional function spaces, rather than functions between finite dimensional Euclidean spaces~\citep{kovachki2023neural}. For problems governed by partial differential equations (PDEs), one learns the \emph{solution operator} mapping a problem input, a forcing term, an initial or boundary condition, or a coefficient field, to the corresponding solution function. Once trained, such an operator predicts the solution for a new input in a single forward pass, without a mesh and without re-solving the equation, so that a single model represents an entire family of PDE instances rather than one~\citep{kovachki2023neural,li2021fno}. This makes operator learning an attractive surrogate for the repeated query and many query settings that pervade computational science and uncertainty quantification.

The theoretical foundation of operator learning is the universal approximation theorem for operators of \citet{chen1995universal}, which established that neural networks can approximate continuous nonlinear operators, through a two network construction that later architectures would adopt. Two such architectures now dominate the field. The Deep Operator Network (DeepONet) \citep{lu2021deeponet} is a direct realization of the Chen and Chen theorem and is the architecture we build on; we describe it in Section~\ref{sec:ol-deeponet}. The Fourier Neural Operator \citep{li2021fno} instead parameterizes the integral kernel in Fourier space and can be viewed as a particular parameterization of the same operator approximation idea. Both are mesh free and discretization flexible, and both inherit the universal approximation property. In this work we adopt DeepONet as the deterministic backbone of our framework.

\subsection{DeepONet}\label{sec:ol-deeponet}
DeepONet~\cite{lu2021deeponet} approximates an operator $\mathcal{G}: \mathcal{V} \to \mathcal{U}$ between two function spaces through two jointly trained subnetworks, known as the branch and trunk networks. The \emph{branch} network reads the input function $v \in \mathcal{V}$, discretized at a fixed set of $m$ sensor locations $\{x_1,\dots,x_m\}$, and outputs a coefficient vector $[b_1(v),\dots,b_p(v)]$. The \emph{trunk} network reads a query location $y$ in the domain of the output function and outputs a basis vector $[t_1(y),\dots,t_p(y)]$. The operator is then evaluated by combining the two through an inner product,
\begin{equation}\label{eq:deeponet}
    \mathcal{G}(v)(y) \;\approx\; \sum_{k=1}^{p} b_k(v)\,t_k(y) \;+\; b_0,
\end{equation}
where $b_0 \in \mathbb{R}$ is a bias. Intuitively, the trunk network learns a data driven basis for the output function while the branch network learns the coefficients of the input function in that basis. This decomposition is precisely the structure underlying the universal approximation theorem for operators \citep{chen1995universal}, which guarantees that an approximation of the form \eqref{eq:deeponet} can represent any continuous operator to arbitrary accuracy given sufficient network capacity.

One advantage of DeepONet is its architectural flexibility, as the branch and trunk networks can employ different neural network architectures, including fully connected and convolutional networks, provided that their outputs have the same latent dimension $p$. This flexibility allows DeepONet to be adapted to the structure of time-dependent problems, as considered in Section~\ref{sec:experiments}. However, the standard DeepONet architecture is deterministic and produces a single output function for each input, which limits its ability to represent the intrinsic randomness of stochastic operators. This limitation motivates the stochastic extension introduced in the following subsection.

\subsection{Stochastic Operator Learning}\label{sec:ol-son}
In many scientific settings, the data are noisy or the underlying operator is stochastic. For example, in stochastic partial differential equations (SPDEs), randomness may arise from stochastic forcing, random coefficients, or uncertain initial conditions. In such settings, deterministic operator approximations provide only point predictions and do not directly characterize the uncertainty in the predicted solutions. Several approaches incorporate uncertainty quantification (UQ) into operator learning through Bayesian neural networks or ensemble-based methods, including Bayesian DeepONet trained using stochastic gradient Langevin dynamics \citep{lin2023bdeeponet}, variational Bayesian DeepONet \citep{garg2023vbdeeponet}, and ensembles constructed using randomized priors \citep{yang2022randomized}. In contrast, the framework considered in this work introduces stochasticity directly into the network dynamics through a stochastic differential equation (SDE).
\subsubsection{Stochastic Neural Networks}
{The stochastic operator learning framework considered in this work builds upon the stochastic neural network (SNN) mechanism, which extends the neural ordinary differential equation (Neural ODE) framework~\cite{Chen2018,Dupont2019,Haber2018} by incorporating stochastic diffusion into the hidden-layer dynamics. This formulation allows the network to generate multiple output realizations and characterize the uncertainty associated with its predictions~\cite{archibald2021bsde}. Specifically, the discrete-layer SNN considered in this work is defined by
\begin{equation}
\label{SNN_1layer}
\begin{array}{l}
A_{n+1} = A_n + h f(A_n,\theta_n) + \sqrt{h}\, g(A_n,\theta_n)\omega_n, \qquad n=0,1,\ldots,N-1,
\end{array}
\end{equation}
where \(A_n \in \mathbb{R}^d\) denotes the state of the network at the \(n\)th layer, \(f\) is the drift term corresponding to a DNN architecture, \(g\) is a matrix-valued diffusion function taking values in \(\mathbb{R}^{d\times r}\) that controls and characterizes the magnitude and structure of the uncertainty in the SNN, \(\theta_n\) denotes the set of parameters at the \(n\)th layer, \(h>0\) is the step-size that can serve as a stabilizing factor for the DNN, and \(\omega_n\) is a standard \(r\)-dimensional Gaussian random vector.

The training of Neural ODEs can be formulated as an optimal control problem, where the network parameters serve as control variables and the dynamics are governed by an ODE. Similarly, the training of the SNN model~\eqref{SNN_1layer} can be formulated as a stochastic optimal control problem.

To this end, let \(T>0\) be a fixed pseudo-terminal time, and consider the limit \(N\to\infty\). In this limit, the discrete dynamics in Eq.~\eqref{SNN_1layer} formally converge to the continuous-time SDE
\begin{equation}
\label{SNN_continuous}
\begin{array}{l}
A_T = A_0 + \mathlarger{\int}_0^T f(A_t,\theta_t)\,dt + \mathlarger{\int}_0^T g(A_t,\theta_t)\,dW_t,
\end{array}
\end{equation}
where \(W=\{W_t\}_{0\le t\le T}\) is a standard Brownian motion associated with the i.i.d. Gaussian sequence \(\{\omega_n\}_n\) in Eq.~\eqref{SNN_1layer}. The time-dependent parameter \(\theta_t\) in Eq.~\eqref{SNN_continuous} is then treated as a control process, and the training problem is formulated as the stochastic optimal control problem
\begin{equation}
\begin{array}{l}
\label{optimal_control}
J(\theta^*)=\inf\limits_{\theta\in\Theta} J(\theta),
\end{array}
\end{equation}
where \(\Theta\) is a convex admissible control set and the cost functional \(J\) is defined by
\begin{align*}
J(\theta)=\mathbb{E}\left[\Phi(A_T,\Gamma)+\mathlarger{\int}_0^T r(A_t,\theta_t)\,dt\right].
\end{align*}
Here, \(\Gamma\) denotes the target random variable associated with the training data, \(\Phi(A_T,\Gamma):=\|A_T-\Gamma\|_{\text{loss}}\) is the terminal loss, and \(\int_0^T r(A_t,\theta_t)\,dt\) is the running cost. 

Two classical approaches to stochastic optimal control are the dynamic programming principle (DPP) and the stochastic maximum principle (SMP). The former leads to the Hamilton-Jacobi-Bellman (HJB) equation~\cite{FlemingRishel1975,FlemingSoner2006, Yong1999}, whose numerical solution becomes computationally challenging in high-dimensional settings. In contrast, the SMP characterizes optimal controls through a Hamiltonian and an associated system of forward-backward stochastic differential equations (SDEs). To reduce the computational cost of this approach, \citet{archibald2020sgdsoc} proposed a stochastic gradient descent method based on sampled trajectories. This sample wise scheme was later extended to batch sampling by \citet{sun2025batchsmp}, who employed a higher-order forward scheme over a batch of trajectories to reduce the variance of the gradient estimator and improve the convergence rate.

To this end, we further reformulate~\eqref{SNN_continuous}-\eqref{optimal_control} as 
an optimization problem characterized through the following Hamiltonian:
\begin{equation}
\label{Hamiltonian_loss}
 H(a, b, c, u) = r(a, u) + b^{\top}f(a, u)
+\mathrm{tr}\left(c^{\top}g(a, u)\right).
\end{equation}
For any admissible control $\theta\in\Theta$ satisfying 
$\mathbb{E}[|\theta|^2]<\infty$, let $A_{t,\theta}$ denote the corresponding state 
process solving Eq.~\eqref{SNN_continuous}. We then consider the associated 
adjoint processes $(B_{t,\theta},C_{t,\theta})$, which satisfy (see \cite{Bao_Control_20, liang2024convergence, Liang_2025_FoDS})
\begin{equation}
\label{adjoint_SDE}
dB_{t,\theta}
=
-\bigl(r_a(A_{t,\theta},\theta_t)
+f_a(A_{t,\theta},\theta_t)^{\top}B_{t,\theta}
+g_a(A_{t,\theta},\theta_t)^{\top}C_{t,\theta}\bigr)\,dt
-C_{t,\theta}\,dW_t,
\end{equation}
with terminal condition
\[
B_{T,\theta}=\Phi_a(A_{T,\theta},\Gamma).
\]
Here, \(r_a\), \(f_a\), \(g_a\), and $\Phi_a$ denote the corresponding derivatives 
with respect to \(a\), and \(C_{t,\theta}\) is the martingale term in the backward 
adjoint equation. The stochastic control problem~\eqref{optimal_control} can then be characterized in terms of the Hamiltonian~\eqref{Hamiltonian_loss} through the necessary optimality condition provided by the stochastic maximum principle (SMP)~\cite{Andersson2009,Sun2023}, as stated in the following theorem.
\begin{theorem}
\label{theorem: SMP}
Assume that \(\theta^*\) is an optimal control for the control 
problem~\eqref{optimal_control}, and let \(A_{t,\theta^*}\) be the solution to 
Eq.~\eqref{SNN_continuous} corresponding to \(\theta^*\). Then, there exist 
adjoint processes \((B_{t,\theta^*},C_{t,\theta^*})\) satisfying the backward 
SDE~\eqref{adjoint_SDE} associated with \(A_{t,\theta^*}\) and \(\theta_t^*\). 
Moreover, \(\theta_t^*\) satisfies the pointwise minimization condition:
\begin{equation}
\label{optimize_Hamiltonian}
H(A_{t,\theta^*},B_{t,\theta^*},C_{t,\theta^*},\theta_t^*)
=
\inf_{\theta\in\Theta}
H(A_{t,\theta^*},B_{t,\theta^*},C_{t,\theta^*},\theta).
\end{equation}
\end{theorem}

Since both the optimal control problem~\eqref{optimal_control} and the Hamiltonian minimization problem~\eqref{optimize_Hamiltonian} are formulated in continuous time, they must be discretized in order to train the SNN model~\eqref{SNN_continuous}. We refer readers to~\citet{archibald2021bsde} for the numerical discretization and implementation details, and to~\citet{archibald2024convergence} for the convergence analysis and error estimates of the sample wise backpropagation scheme.
}

\subsubsection{Stochastic operator networks}
{In this section, we briefly discuss how to extend SNN to a stochastic operator learning framework by utilizing the architecture of DeepONet. This extension leads to the Stochastic Operator Network (SON), which
will enable probabilistic prediction for operator learning tasks. 

A straightforward approach is to formulate the branch network in the DeepONet by an SNN governed by Eq.~\eqref{SNN_1layer}, with the initial state \(A_0\) given by the discretization of the input function \(v\) at the sensor locations. The branch network is then defined by the final state of the SDE~\eqref{SNN_1layer} after $N$ pseudo-time steps:
\[
\beta(u;\theta_\beta,\epsilon):=A_N(u,\epsilon;\theta_\beta),
\]
where $\theta_\beta=\{\theta_n\}_{n=0}^{N-1}$ collects all trainable parameters associated with the drift and diffusion terms.

The trunk network can be represented by a standard feed-forward or convolutional neural network, with its output dimension matching that of the stochastic branch network and is denoted by
\[
\tau(\pmb{y};\theta_\tau),
\]
where $\theta_\tau$ collects all trainable parameters of the trunk network. We then combine these two outputs to form the SON operator as in~\eqref{eq:deeponet} to obtain:
\begin{equation}
\label{StoDeepONet}
\hat{G}_{\text{SON}}(u)(y,\epsilon)
=
\pmb{d}\bigl(\beta(u;\theta_\beta,\epsilon),\tau(\pmb{y};\theta_\tau)\bigr)+b_0.
\end{equation}
This formulation leads to the original SON developed in~\cite{bausback2025son}. However, for this approach, one need to modify the Hamiltonian introduced in~\eqref{Hamiltonian_loss} due to the presence of extra parameters in the trunk network~\cite{huynh2026diffusion}. Moreover, the training the original SON requires the updates of the parameters of the branch network, the diffusion term and the trunk network within a single training loop. This can be computationally expensive when the target system is complex and requires large branch and trunk networks. 

To address this issue, a subsequent \emph{two-phase} training procedure for the SON was introduced in~\cite{huynh2026diffusion} which consists of a pre-training phase (Phase I) and a fine-tuning phase (Phase II).  In Phase I, we train a deterministic neural operator based on DeepONet, denoted by $S_{\text{DO}}$, using the noisy training data to obtain a baseline solver. In Phase II, we apply the SNN to the output of $S_{\text{DO}}$ and propagate this output through the SDE dynamics~\eqref{SNN_1layer} to produce a probabilistic prediction. Since the deterministic operator captures the overall solution structure, the drift term in Phase II is primarily responsible for maintaining or refining the baseline prediction, while the diffusion term characterizes the uncertainty. Consequently, the drift can be represented by a relatively simple function rather than a large neural network, allowing the second phase to focus on learning the stochastic variations. Furthermore, since the SNN is applied directly to the output of $S_{\text{DO}}$, the trunk network does not need to be incorporated into the SNN layers. Consequently, the Hamiltonian formulation in~\eqref{Hamiltonian_loss} can be used directly for training Phase II.

\begin{remark}
An additional consideration is the calibration of the
diffusion magnitude. Since the Hamiltonian-based training
does not directly determine the overall magnitude of
the stochastic fluctuations, this magnitude is calibrated
using the residual between the noisy observations and
the deterministic prediction. Specifically, the target
noise level is estimated as
\begin{equation}\label{eq:target-sigma}
\sigma_\star
= \operatorname{std}
\big(u-S^{\mathrm{DO}}(v)\big),
\end{equation}
where $u$ denotes the noisy solution data associated with the input function $v$. A scalar magnitude parameter is then selected so that the predicted stochastic fluctuations match the estimated noise level. This parameter is fixed during training, allowing the diffusion network to focus on learning the structure of the uncertainty. 
\end{remark}

The separation of deterministic and stochastic learning in the two-phase SON also facilitates its extension to federated learning, as the two phases can be trained separately under a federated protocol. Building upon this formulation~\citep{huynh2026diffusion}, we introduce the \textit{Federated stochastic operator network} (FedSON) framework in next section, where federated training is applied to both phases.
}

\section{Federated Stochastic Operator Network}\label{sec:fedson}

{
Building upon the two-phase SON formulation, FedSON introduces two additional components: a federated training protocol for both the deterministic and stochastic phases, and an auxiliary data mechanism designed to address data heterogeneity across clients. In particular, the auxiliary data mechanism enables collaborative training when clients possess data drawn from disjoint regions of the input function space, without requiring them to share their original solution data.
}
\subsection{Federated problem setup}\label{sec:fedson-setup}
{
We consider $K$ clients that jointly learn a solution operator without sharing their solution data, that is, the computed PDE solution trajectories that form the training targets. Each client $k$ holds a local dataset $\mathcal{D}_k$ whose inputs are drawn from a client-specific subset $\mathcal{V}_k$ of the input function space $\mathcal{V}$. The federated objective is the size-weighted risk introduced in Section~\ref{sec:fl-fedavg},
\begin{equation}\label{eq:fedson-objective}
\min_{\theta}\;
\mathcal{L}(\theta) \;=\; \sum_{k=1}^{K}\frac{N_k}{N}\,\mathcal{L}_k(\theta),
\qquad
\mathcal{L}_k(\theta) \;=\; \frac{1}{N_k}\!\!\sum_{(v,u)\in\mathcal{D}_k}\!\!
\ell\big(\mathcal{G}_\theta(v),\,u\big),
\end{equation}
where $N_k=|\mathcal{D}_k|$, $N=\sum_k N_k$, and $\mathcal{G}_\theta$ is the operator model. Training proceeds in communication rounds of local updates followed by the federated average of Eq.~\eqref{eq:fl-fedavg}.

The heterogeneity we study is the extreme case in which clients generate input functions using \emph{disjoint subsets of basis functions}. Specifically, we consider the input function space
\begin{equation}\label{eq:input-space}
\mathcal{V} = \operatorname{span}\{T_0(\hat{x}),T_1(\hat{x}),\ldots,T_M(\hat{x})\},
\end{equation}
where $T_i$ denotes the Chebyshev polynomial of the first kind of degree $i$, with the transformed coordinate $\hat{x}=2x-1$ for $x\in[0,1]$. We partition the basis indices into disjoint blocks, one per client, such that
\begin{equation}\label{eq:disjoint}
\mathcal{I}_k \cap \mathcal{I}_j = \varnothing
\quad\text{for all}\quad k\neq j.
\end{equation}
The input functions for client $k$ are generated as
\begin{equation}\label{eq:cheb-partition}
    \mathcal{V}_k \;=\; \Big\{\, v(x) = \textstyle\sum_{i\in \mathcal{I}_k} a_i\,T_i(\hat{x})
    \;:\; a_i \in [-1,1] \,\Big\},
    \qquad a_i \sim \mathcal{U}(-1,1),
\end{equation}
where the index sets $\{\mathcal{I}_k\}_{k=1}^{K}$ are selected from the retained Chebyshev indices and $\mathcal{U}(a,b)$ is the uniform distribution on the interval $[a,b]$ with $a\leq b$. Consequently, each client generates input functions using a distinct set of Chebyshev modes, resulting in heterogeneous local input distributions. As discussed in Section~\ref{sec:fl-hetero}, such heterogeneity can adversely affect the performance of federated averaging. Addressing the challenges associated with such heterogeneous data partitions is a primary objective of FedSON.
}
\vspace{-0.2cm}

\subsection{Phase I: federated deterministic drift}\label{sec:fedson-phase1}

{The first phase learns the deterministic drift, which approximates the mean solution operator. Any suitable DeepONet architecture can be employed in this phase, and the specific architectures used in our experiments are described in Section~\ref{sec:experiments}.} The drift network $\mathcal{G}^{\mathrm{drift}}_{\theta}$ maps an input function $v$ to the predicted mean solution $\bar{u}=\mathcal{G}^{\mathrm{drift}}_{\theta}(v)$, and is trained on the (noisy) local data under the mean squared error
\begin{equation}\label{eq:phase1-loss}
    \mathcal{L}_k^{\mathrm{I}}(\theta)
    \;=\; \frac{1}{N_k}\sum_{(v,u)\in\mathcal{D}_k}
    \big\| \mathcal{G}^{\mathrm{drift}}_{\theta}(v) - u \big\|_2^2 .
\end{equation}
Each client performs several local epochs of stochastic gradient descent on Eq.~\eqref{eq:phase1-loss}, after which the server aggregates the client models by the size weighted average of Eq.~\eqref{eq:fl-fedavg}. Because the drift is trained on noisy data under a mean squared loss, it recovers the conditional mean of the solution and leaves the fluctuations of the data about that mean to the second phase. At the end of Phase~I the aggregated drift is frozen; it is not updated again.

{When trained using standard FedAvg under the disjoint data partition of Eq.~\eqref{eq:disjoint}, the deterministic operator may experience performance degradation due to differences between the local training distributions. Such heterogeneity can increase discrepancies between local model updates and adversely affect the performance of the aggregated model, as discussed in Section~\ref{sec:fl-hetero}. To address this challenge, we introduce an auxiliary data mechanism in the following subsection.
}

\subsection{Auxiliary data through pseudolabel exchange}\label{sec:fedson-aux}

{The core idea of FedSON is to introduce auxiliary data that allow clients to train on overlapping input distributions without sharing their original solution data. Specifically, each client uses a locally trained model to generate predictions for its input functions and shares the resulting input-prediction pairs with the other clients. These predictions are referred to as pseudolabels, and the corresponding input-prediction pairs are used to augment the local training datasets. The construction proceeds in three steps.}

{First, each client $k$ trains a local drift model using its own dataset $\mathcal{D}_k$, producing a model $\mathcal{G}^{(k)}$ that approximates the solution operator on the local input subspace $\mathcal{V}_k$. This initial training stage uses the same loss function as Eq.~\eqref{eq:phase1-loss} and requires no communication between clients.

Next, clients exchange their locally trained models and input functions $v\in\mathcal{V}_k$. Upon receiving another client's model and input functions, each client evaluates the received model on the corresponding inputs to generate pseudolabels. Specifically, for each pair of clients $k\neq j$, client $k$ obtains the following auxiliary dataset from client $j$:
\begin{equation}\label{eq:pseudolabels}
\mathcal{A}_{k\leftarrow j}
\;=\; \Big\{\, \big(v,\; \mathcal{G}^{(j)}(v)\big)
\;:\; v \in \mathcal{V}_j \,\Big\},
\end{equation}
in which the inputs $v$ are client $j$'s own input functions and the labels are the teacher predictions $\mathcal{G}^{(j)}(v)$ rather than ground truth solutions. Only the model parameters and the input functions are communicated; no solution trajectory, neither the noisy data nor the clean trajectories, ever leaves any client.

Finally, each client constructs an augmented dataset by combining its original training data with the pseudolabeled data obtained from the other clients:
\begin{equation}\label{eq:aug-dataset}
    \widetilde{\mathcal{D}}_k
    \;=\; \mathcal{D}_k \;\cup\; \bigcup_{j\neq k}\mathcal{A}_{k\leftarrow j} .
\end{equation}
Whereas $\mathcal{D}_k$ covers only $\mathcal{V}_k$, the augmented dataset $\widetilde{\mathcal{D}}_k$ covers the full function space $\mathcal{V}=\bigcup_j \mathcal{V}_j$, and it does so identically for every client. The disjoint supports of Eq.~\eqref{eq:disjoint} are thereby replaced by overlapping, effectively homogeneous supports. Federated training in Phase~I is then carried out on the augmented datasets, replacing $\mathcal{D}_k$ by $\widetilde{\mathcal{D}}_k$ in Eq.~\eqref{eq:phase1-loss}. Because the clients now optimize over a common region of function space, the local models are compatible and the federated average of Eq.~\eqref{eq:fl-fedavg} combines rather than degrades them, which directly addresses the weight divergence problem of Section~\ref{sec:fl-hetero}. The global model is initialized with a new set of random parameters rather than by averaging the locally trained DeepONets.
}

\subsection{Phase II: federated stochastic diffusion}\label{sec:fedson-phase2}
{The second phase learns the stochastic uncertainty of the solution around the deterministic prediction obtained in Phase~I. Following the two-phase SON formulation of \citet{huynh2026diffusion}, the deterministic operator is frozen, and its output $\bar{u}=\mathcal{G}^{\mathrm{drift}}(v)$ serves as the initial state of the SNN. The stochastic evolution and training procedure follow the formulation introduced in Section~\ref{sec:ol-son}.

In FedSON, each client trains the diffusion model locally using the stochastic maximum principle, while the server aggregates the updated parameters according to FedAvg~\eqref{eq:fl-fedavg}. The training is performed using the augmented datasets $\widetilde{\mathcal{D}}_k$ constructed in Section~\ref{sec:fedson-aux}. The complete FedSON training procedure is summarized in Algorithm~\ref{alg:fedson}.
}

\begin{algorithm}[h!]
\caption{FedSON training}\label{alg:fedson}
\begin{algorithmic}[1]
\State \textbf{Input:} client datasets $\{\mathcal{D}_k\}$, index sets $\{\mathcal{I}_k\}$.
\State Each client $k$ trains a local DeepONet
$\mathcal{G}^{(k)}$ using $\mathcal{D}_k$.
\State Clients exchange their locally trained models and input functions
to construct the augmented datasets $\widetilde{\mathcal{D}}_k$
using Eqs.~\eqref{eq:pseudolabels}--\eqref{eq:aug-dataset}.
\Statex \textbf{Phase I: Federated deterministic training}.
\State Initialize a new global DeepONet with random parameters $\theta^0$.
\For{round $r=0,1,\dots$}
\State each client updates $\theta^r$ locally on $\widetilde{\mathcal{D}}_k$ via Eq.~\eqref{eq:phase1-loss}
\State server aggregates: $\theta^{r+1}\leftarrow\sum_k \tfrac{N_k}{N}\theta^{r}_k$
\EndFor
\State freeze the drift $\mathcal{G}^{\mathrm{drift}}$
\Statex \textbf{Phase II: stochastic diffusion}.
\State calibrate the noise magnitude to $\sigma_\star$ by Eq.~\eqref{eq:target-sigma}
\For{round $r=0,1,\dots$}
\State each client updates the diffusion locally on $\widetilde{\mathcal{D}}_k$
by the SMP update Eq.~\eqref{eq:smp-update}
\State server aggregates the diffusion parameters by Eq.~\eqref{eq:fl-fedavg}
\EndFor
\State \textbf{Output:} federated drift $\mathcal{G}^{\mathrm{drift}}$ and diffusion $\sigma_{\theta_g}$
\end{algorithmic}
\end{algorithm}
{In Section~\ref{sec:experiments}, we evaluate FedSON on three PDE problems and compare its performance with heterogeneous and homogeneous federated baselines, as well as a centralized model.}

\section{Experiments and Results} \label{sec:experiments}
{
We evaluate FedSON under four training settings. In the \emph{heterogeneous setting}, each client trains using its original dataset drawn from a distinct input distribution. In the \emph{homogeneous setting}, the original training data are pooled and redistributed across clients to obtain approximately independent and identically distributed (IID) local datasets, followed by standard FedAvg training. In the \emph{auxiliary setting}, each client's original dataset is augmented with pseudolabeled samples from the other clients, following the procedure described in Section~\ref{sec:fedson-aux}. Finally, the \emph{centralized setting} trains a single model directly on the pooled original datasets without federated aggregation. One can find the code \href{https://github.com/TonyTangTT/FedSON}{here}. 
}
\subsection{Heat Equation}\label{sec:heat}

We first consider a one dimensional stochastic heat equation with a client dependent forcing term. This experiment is designed to test the federated pipeline in a setting where each client observes the same physics but is driven by a distinct, non-overlapping family of source functions, so that no single client sees the full input space during local training.

\subsubsection{Governing Equation}

Let $u(x,t)$ denote the temperature field on the unit interval $x \in [0,1]$ over the time horizon $t \in [0,T]$ with $T = 1$. The field evolves according to the forced heat equation
\begin{equation}\label{eq:heat-pde}
\frac{\partial u}{\partial t}
\;=\; \kappa\,\frac{\partial^2 u}{\partial x^2} \;+\; 5\, f(x),
\qquad x \in (0,1),\; t \in (0,T],
\end{equation}
with thermal diffusivity $\kappa = 1$. The forcing term $f(x)$ is time-independent and is the quantity that varies across clients; its construction is detailed in Section~\ref{sec:heat-forcing}. The fixed gain of $5$ increases the magnitude of the deterministic response relative to the additive noise; the networks are supplied with the unscaled field $f$ and internalize this constant.

We impose inhomogeneous Dirichlet boundary conditions
\begin{equation}\label{eq:heat-bc}
u(0,t) = a_0 = 0, \qquad u(1,t) = a_1 = 1, \qquad t \in [0,T],
\end{equation}
and take the initial condition to be the linear profile that is consistent with the boundary data,
\begin{equation}\label{eq:heat-ic}
    u(x,0) = x, \qquad x \in [0,1].
\end{equation}
The initial condition~\eqref{eq:heat-ic} is fixed and shared across all clients and all samples; the only quantity that distinguishes one trajectory from another is the forcing $f$.

To model observational and process noise, the numerically integrated trajectory is perturbed by additive Gaussian noise at every time step, so that the data seen by the network are realizations of a stochastic process rather than the deterministic solution of~\eqref{eq:heat-pde}. This stochasticity is what the second-stage stochastic model is ultimately trained to capture.

\subsubsection{Spatial and Temporal Discretization}

{
The spatial domain $[0,1]$ is discretized using $N_x=100$ uniform subintervals, while the time horizon $[0,T]$ is divided into $N_t=80$ uniform steps. Equation~\eqref{eq:heat-pde} is discretized using the backward Euler method in time and second-order central finite differences in space. The resulting tridiagonal linear systems are solved using the Thomas algorithm. Each solution trajectory consists of $80$ snapshots, each containing $101$ spatial values.

To generate stochastic trajectories, a Gaussian perturbation is introduced at each time step:
\begin{equation}\label{eq:heat-noise}
\tilde{u}^{\,i+1}
= u^{i+1} + \sigma\sqrt{\Delta t}\,\xi^{i+1},
\qquad \xi^{i+1}\sim\mathcal{N}(0,I),
\end{equation}
with $\sigma=0.35$. The perturbed state is used as the initial state for the subsequent time step, allowing the stochastic fluctuations to propagate throughout the trajectory. Only the resulting noisy
trajectories are used for training. The problem and discretization settings are summarized in Table~\ref{tab:heat-settings}.
}

\begin{table}[h!]
\centering
\caption{Problem and discretization settings for the heat equation experiment.}
\label{tab:heat-settings}
\begin{tabular}{ll}
\hline
Quantity & Value \\
\hline
Domain & $x \in [0,1]$, $t \in [0,1]$ \\
Diffusivity $\kappa$ & $1$ \\
Boundary conditions & $u(0,t)=0,\; u(1,t)=1$ \\
Initial condition & $u(x,0)=x$ \\
Spatial nodes $N_x+1$ & $101$ (uniform, $h=1/100$) \\
Time steps $N_t$ & $80$ ($\Delta t = 1/80$) \\
Time integrator & backward Euler + central differences \\
Noise level $\sigma$ & $0.35$ (added as $\sigma\sqrt{\Delta t}\,\xi$ per step) \\
Forcing amplification & $5$ \\
Chebyshev modes $M+1$ & $30$ (first kind, $T_0 \dots T_{29}$) \\
Clients $C$ & $10$ \\
Modes per client $n$ & $3$ (bands $\{3i,3i+1,3i+2\}$) \\
Coefficient prior & $\mathcal{U}(-1,1)$ on active modes \\
Samples per client & $1000$ ($600$ train / $400$ test) \\
\hline
\end{tabular}
\end{table}

\subsubsection{Client Dependent Forcing}\label{sec:heat-forcing}

{
The federated setup comprises $C=10$ clients. For this experiment, we retain $M+1=30$ Chebyshev modes, from $T_0$ through $T_{29}$, and assign each client a distinct contiguous block of $n=3$ modes. Specifically, for client $i\in\{0,\dots,C-1\}$, the index set is
\begin{equation}\label{eq:client-band}
\mathcal{I}_i
= \{in,in+1,\dots,in+n-1\}
= \{3i,3i+1,3i+2\}.
\end{equation}
Thus, client $0$ is assigned modes $\{0,1,2\}$, client $1$ is assigned modes $\{3,4,5\}$, and so on up to client $9$, which is assigned modes $\{27,28,29\}$. Since $C \times n=10\times3=30=M+1$, the client index sets form a partition of the full set of retained Chebyshev modes. For each client, the active coefficients are drawn independently from $\mathcal{U}(-1,1)$, while all remaining coefficients are set to zero. Coefficients are generated with a fixed client-specific random seed for reproducibility.

For each client, we draw $N=1000$ forcing samples and split them into $600$ training and $400$ test samples. Each forcing sample is used to generate a noisy solution trajectory using the numerical scheme described above. Consistent with the gain in Eq.~\eqref{eq:heat-pde}, the numerical solver integrates the PDE with the forcing $5f$ to increase the magnitude of the deterministic solution relative to the stochastic perturbations; the networks are fed the unscaled field $f$ and internalize the constant. The resulting per-client tensors are the forcing fields $f\in\mathbb{R}^{N\times(N_x+1)}$ and the noisy trajectories $\tilde{u}\in\mathbb{R}^{N\times N_t\times(N_x+1)}$, stored separately for training and testing.
}

\subsubsection{Model Architecture}\label{sec:heat-arch}
The deterministic component of the model is a time marching DeepONet that maps a forcing field to the full space--time trajectory. A \emph{branch} network reads the forcing $f$, and a \emph{trunk} network reads the spatial grid; both are five hidden layer fully connected networks (hidden width $80$, GELU activations, layer normalization, light dropout) that produce a shared latent code of dimension $64$. The branch and trunk codes are combined and decoded into an initial latent state, which is then advanced by a \emph{time-aware propagator}: a residual network that, given the current latent state and a normalized step index $t_{\mathrm{idx}} \in (0,1]$, produces the increment to the next latent state. Rolling the propagator for $N_t$ steps yields a latent trajectory, which a final decoder maps back to physical space. Concretely, the latent state is rolled by the propagator $\mathcal{P}$,
\begin{equation}\label{eq:heat-propagator}
z_0 \;=\; \mathrm{Deco}\big(\mathrm{branch}(f),\, \mathrm{trunk}(\mathrm{grid})\big),
\qquad
z_{k+1} \;=\; z_k \;+\; \mathcal{P}\big(z_k,\, t_{\mathrm{idx}}\big),
\quad t_{\mathrm{idx}} = \tfrac{k+1}{N_t},
\end{equation}
and each latent state is decoded as a residual from the fixed initial condition,
\begin{equation}\label{eq:don-residual}
u_{\mathrm{DON}}(\cdot, t_k) \;=\; u(\cdot, 0) \;+\;
\mathrm{dec}\big(z_k\big).
\end{equation}
The final decoder is zero initialized, so that at the start of training the prediction is exactly the initial profile~\eqref{eq:heat-ic} broadcast across time. This provides the correct inductive bias for a solution that drifts away from a known initial state.

The stochastic component is a Stochastic Operator Network (SON) that wraps the trained DeepONet and models the distribution of trajectories around the deterministic mean.

\subsubsection{Training Overview}

Training proceeds in two phases. In the first phase we fit the deterministic DeepONet (DON) to the mean behavior of the trajectories; in the second phase we freeze the DON and train the SON to capture the fluctuations. The first phase itself is carried out in three stages: a per client warmup that produces one specialized model per client, an auxiliary data construction step that exchanges knowledge between clients, and a federated training run on the combined data. The four stages are summarized in Table~\ref{tab:heat-training} and detailed below. Throughout, the deterministic solution is never used; both phases are supervised only by the noisy trajectories, and clean trajectories are reserved for diagnostics.

Local optimization in all stages uses AdamW with gradient norm clipping, and client models are combined by weighted federated averaging (FedAvg), in which each client's parameters are averaged with weights proportional to its local dataset size.

\begin{table}[h!]
\centering
\caption{[1D Heat Eq.] Summary of the four training stages for the heat equation experiment. All stages use AdamW with gradient clipping; federated stages aggregate by size-weighted FedAvg. ``Rnds.'' is communication rounds and ``Ep.'' is local epochs per round; ``LR start'' and ``LR end'' give the cosine-annealed learning rate range.}
\label{tab:heat-training}
\setlength{\tabcolsep}{4pt}
\begin{tabular}{@{}llccccc@{}}
\hline
Stage & Data & Fed. & Rnds. & Ep. & LR start & LR end \\
\hline
1. Warmup      & own noisy ($600$)    & no  & $30$ & $50$ & $5\text{e-}4$ & $1\text{e-}4$ \\
2. Aux.\ data  & pseudolabels         & --- & ---  & ---  & ---           & ---           \\
3. Fed.\ DON   & expanded ($6000$)    & yes & $30$ & $50$ & $2\text{e-}4$ & $1\text{e-}5$ \\
4. Fed.\ SON   & own noisy ($600$)    & yes & $5$  & $4$  & $1\text{e-}5$ & $1\text{e-}5$ \\
\hline
\end{tabular}
\end{table}

{
The FedSON training procedure follows Algorithm~\ref{alg:fedson}, with the training parameters summarized in Table~\ref{tab:heat-training}. First, each client independently trains a local DeepONet using its $600$ noisy training pairs $(f,\tilde{u})$. The training uses the mean
squared error loss, with $30$ training cycles of $50$ local epochs each. The resulting models are then used to generate pseudolabels for auxiliary data construction, as described in Section~\ref{sec:fedson-aux}. Each client obtains an augmented dataset containing $6000$ input-output pairs, consisting of $600$ original noisy samples and $5400$ pseudolabeled samples from
the other clients.

Next, a new global DeepONet is initialized and trained using FedAvg on the augmented datasets. Each client performs $50$ local epochs per communication round, followed by parameter aggregation at the server. The training consists of $30$ communication rounds, and the best-performing checkpoint is selected based on the held-out test data. The resulting federated DeepONet provides the deterministic prediction $u_{\mathrm{DON}}$ for the second phase.

In Phase~II, the federated DeepONet is frozen, and the stochastic model is trained to characterize the uncertainty around its deterministic prediction. Specifically, the stochastic state evolves over a pseudo-time interval $\tau\in[0,1]$ using $N_{\ell}=20$ Euler-Maruyama steps of size $h=1/N_{\ell}$,
\begin{equation}\label{eq:son-sde-ou}
u_{n+1}= u_n-\lambda(u_n-u_{\mathrm{DON}})h
+s\,\Sigma(u_n)\sqrt{h}\,\xi_n,
\qquad \xi_n\sim\mathcal{N}(0,I),
\end{equation}
where $\lambda=10^{-3}$ is the mean-reversion coefficient, $\Sigma$ is a learnable low-rank diffusion operator constructed from $r=5$ diffusion networks, and $s=\exp(\text{log-scale})$ controls the diffusion magnitude. The diffusion model is trained using the stochastic maximum principle and the associated backward SDE formulation described in Section~\ref{sec:ol-son}. At each gradient update, the Hamiltonian
is averaged over $8$ independent noise realizations to reduce the variance of the gradient estimate. The diffusion parameters
are aggregated using FedAvg.
}

Before training, the diffusion scale is calibrated to the data: the target noise level is measured as the standard deviation of the residual between the noisy training trajectories and the DeepONet mean, and the log-scale parameter is shifted so that the SON's terminal spread matches this level at initialization. The scale is then held fixed during training while the diffusion heads learn the spatial and temporal \emph{shape} of the fluctuations. Training is federated in the same manner as Stage~3: $5$ communication rounds, $4$ local epochs per client per round, learning rate $10^{-5}$, with FedAvg aggregation across the ten clients. At inference, the trained SON is rolled out repeatedly with fresh noise to produce an ensemble of trajectories whose empirical statistics form the predicted uncertainty envelope around the DeepONet mean.

\subsubsection{Deterministic Operator Results}\label{sec:heat-don-results}

{
We evaluate the deterministic DeepONet under three training regimes: heterogeneous FedAvg, homogeneous FedAvg, and the proposed auxiliary approach. All three use the same model architecture and training configuration, with differences in their training data. The homogeneous setting redistributes
the original training samples across clients to provide a reference with approximately homogeneous data distributions, whereas the auxiliary approach uses pseudolabeled samples without exchanging the original solution trajectories.

Table~\ref{tab:heat-don} reports the per-client relative $L_2$ errors and MSEs against the noisy test trajectories, together with the diagnostic $\lVert u-x\rVert_{\mathrm{rel}}$, which measures the departure from the initial profile $u(x,0)=x$. The heterogeneous FedAvg error is particularly large for client~$0$, whose trajectories exhibit the greatest departure from the initial profile. In contrast,
the errors decrease substantially for clients associated with higher-frequency forcing modes, where the solution remains closer to the initial profile.

The auxiliary approach significantly improves prediction accuracy, particularly for the low-frequency clients. For client~$0$, the relative $L_2$ error decreases from $0.432$ under heterogeneous FedAvg to $0.081$ with auxiliary training, compared with $0.070$ for homogeneous
FedAvg. Across all clients, the auxiliary approach achieves a mean relative $L_2$ error of $0.071$, close to the homogeneous reference value of $0.070$. These results demonstrate that auxiliary data can substantially mitigate the effects of heterogeneous client distributions without sharing the original solution trajectories.
}

\begin{table}[h!]
\centering
\caption{[1D Heat Eq.] Per-client relative $L_2$ errors and MSEs of the three DeepONet training regimes for the heat equation, evaluated against noisy test trajectories. The quantity $\lVert u-x\rVert_{\mathrm{rel}}$ measures the relative departure from the initial profile $u(x,0)=x$.}
\label{tab:heat-don}
\setlength{\tabcolsep}{3pt}
\begin{tabular}{@{}c c c c c c c c@{}}
\hline
 & & \multicolumn{2}{c}{Heterogeneous} & \multicolumn{2}{c}{Homogeneous (reference)} & \multicolumn{2}{c}{Auxiliary (ours)} \\
\cline{3-4}\cline{5-6}\cline{7-8}
Client & $\lVert u-x\rVert_{\mathrm{rel}}$ & rel $L_2$ & MSE & rel $L_2$ & MSE & rel $L_2$ & MSE \\
\hline
$0$ & $0.448$ & $0.432$ & $7.87\text{e-}2$ & $0.070$ & $2.08\text{e-}3$ & $\mathbf{0.081}$ & $2.79\text{e-}3$ \\
$1$ & $0.126$ & $0.119$ & $4.82\text{e-}3$ & $0.073$ & $1.84\text{e-}3$ & $\mathbf{0.072}$ & $1.77\text{e-}3$ \\
$2$ & $0.074$ & $0.075$ & $1.89\text{e-}3$ & $0.070$ & $1.64\text{e-}3$ & $\mathbf{0.070}$ & $1.65\text{e-}3$ \\
$3$ & $0.070$ & $0.071$ & $1.70\text{e-}3$ & $0.070$ & $1.63\text{e-}3$ & $\mathbf{0.069}$ & $1.62\text{e-}3$ \\
$4$ & $0.070$ & $0.071$ & $1.68\text{e-}3$ & $0.069$ & $1.62\text{e-}3$ & $\mathbf{0.069}$ & $1.62\text{e-}3$ \\
$5$ & $0.069$ & $0.071$ & $1.68\text{e-}3$ & $0.069$ & $1.62\text{e-}3$ & $\mathbf{0.069}$ & $1.62\text{e-}3$ \\
$6$ & $0.069$ & $0.071$ & $1.68\text{e-}3$ & $0.069$ & $1.61\text{e-}3$ & $\mathbf{0.069}$ & $1.61\text{e-}3$ \\
$7$ & $0.069$ & $0.070$ & $1.67\text{e-}3$ & $0.069$ & $1.61\text{e-}3$ & $\mathbf{0.069}$ & $1.61\text{e-}3$ \\
$8$ & $0.069$ & $0.070$ & $1.66\text{e-}3$ & $0.069$ & $1.61\text{e-}3$ & $\mathbf{0.069}$ & $1.61\text{e-}3$ \\
$9$ & $0.069$ & $0.070$ & $1.66\text{e-}3$ & $0.069$ & $1.61\text{e-}3$ & $\mathbf{0.069}$ & $1.61\text{e-}3$ \\
\hline
mean  & --- & $0.112$ & $9.72\text{e-}3$ & $0.070$ & $1.69\text{e-}3$ & $0.071$ & $1.75\text{e-}3$ \\
worst & --- & $0.432$ & $7.87\text{e-}2$ & $0.073$ & $2.08\text{e-}3$ & $0.081$ & $2.79\text{e-}3$ \\
\hline
\end{tabular}
\end{table}
\vspace{-0.3cm}
\begin{figure}[h!]
\centering
\begin{minipage}{\textwidth}
\includegraphics[width=0.9\textwidth]{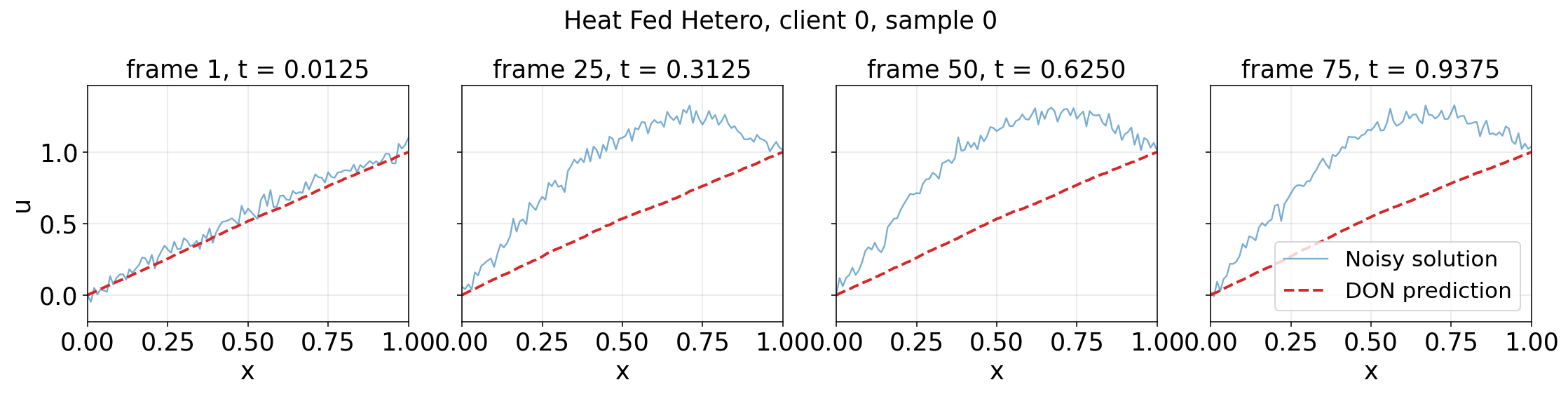}
\end{minipage}

\begin{minipage}{\textwidth}
\includegraphics[width=0.9\textwidth]{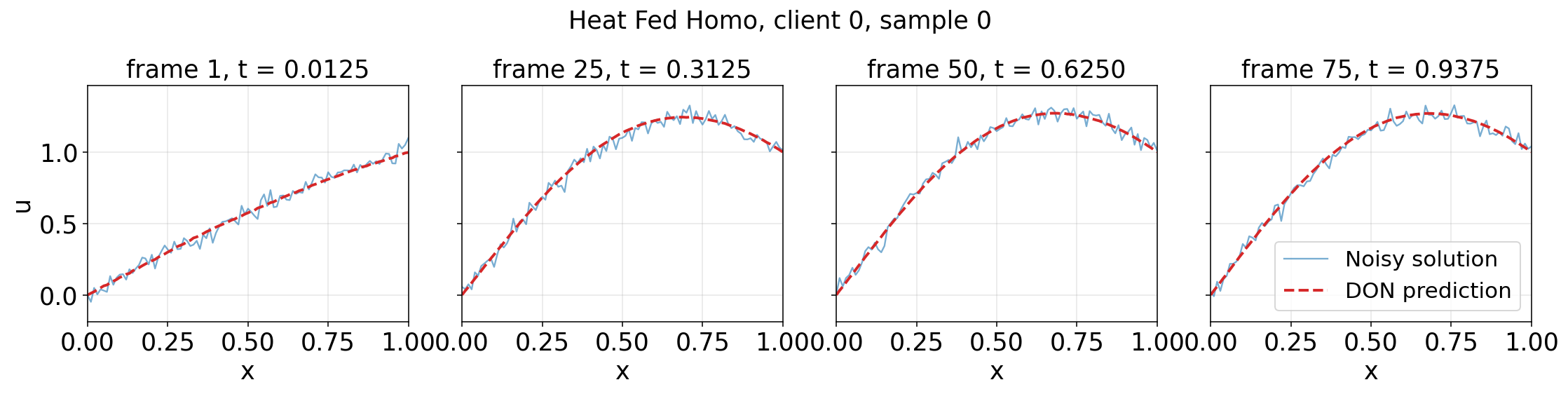}
\end{minipage}

\begin{minipage}{\textwidth}
\includegraphics[width=0.9\textwidth]{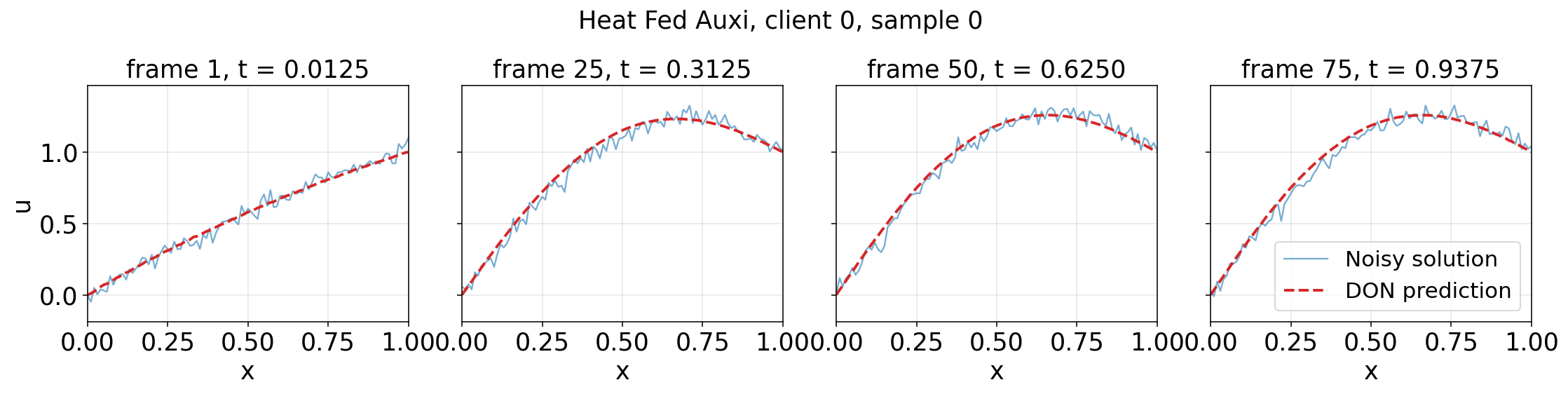}
\end{minipage}
\caption{[1D Heat Eq.] Comparison of predicted spatial profiles (dashed red) and noisy data (solid blue) at four time levels for client~$0$, test sample~$0$. From top to bottom: heterogeneous FedAvg, homogeneous FedAvg, and the auxiliary approach.}
\label{fig:heat-don-flat}
\end{figure}

Figure~\ref{fig:heat-don-flat} compares the predicted
spatial profiles at four time levels for a representative test trajectory from client~$0$. Under heterogeneous FedAvg (top row), the prediction remains close to the initial profile $u\approx x$ and fails to capture the solution structure. In contrast, homogeneous FedAvg (middle row)
and the auxiliary approach (bottom row) closely follow the noisy trajectories.



\subsection{Burgers' Equation} \label{sec:burgers}
We next investigate whether the proposed FedSON framework remains effective for nonlinear dynamical systems. In particular, we consider the viscous Burgers' equation to evaluate the effectiveness of auxiliary data in federated operator learning under heterogeneous client distributions and to assess the stochastic model's ability to characterize uncertainty in the predicted solutions.

The viscous Burgers' equation we consider in this work is given by
\begin{equation}\label{eqn:burgers}
\frac{\partial u}{\partial t} + u \frac{\partial u}{\partial x}
= \nu \frac{\partial^2 u}{\partial x^2}, \; x \in [-1, 1], \; t \in [0, 0.25],
\end{equation} 
with a diffusion coefficient $\nu = 1.0$ and periodic boundary conditions. The spatial and temporal discretizations are set to $\Delta{x} = 0.01$ and $\Delta{t} = 1/3000$, respectively. The initial condition for each client is constructed as a linear combination of Chebyshev polynomials of the first kind, following the construction used for the heat equation. Specifically, client $k$ uses the Chebyshev modes $\mathcal{I}_k=\{3k,3k+1,3k+2\}$, $k=0,\ldots,4$,
with coefficients independently sampled from $\mathcal{U}(-1,1)$ and all remaining coefficients set to zero. To render the initial condition consistent with the periodic boundary conditions, we apply the linear shift
\begin{equation}\label{eqn:burgers_IC}
g_k(x) = f_k(x) - f_k(a) - \frac{f_k(b) - f_k(a)}{b - a} \cdot (x - a), 
\end{equation}
where $f_k$ denotes the raw Chebyshev combination for client $k$ and $a = -1$, $b = 1$ are the endpoints of the spatial domain. The shifted initial condition satisfies $g_k(a) = g_k(b) = 0$. For $t > 0$ the boundary values remain equal to one another but need not vanish; that is, $u(a, t) = u(b, t)$ for all $t \in [0, 0.25]$, with $u(a, 0) = u(b, 0) = 0$ holding only at the initial time.

\subsubsection{Data Generation} \label{sec:burgers_data_generation}
The equation is solved numerically for each client using its initial condition $g_k(x)$ from \eqref{eqn:burgers_IC}. Spatial derivatives are computed with a Fourier pseudospectral method, and time integration uses an implicit–explicit SSPRK2 scheme, with the diffusion term treated implicitly and the advection term explicitly. The solution $u$ is recorded every ten steps; since $T = 0.25$ and $\Delta{t} = 1/3000$, this yields $75$ saved time steps per initial condition. For each client, we generate $1000$ initial conditions, partitioned into $600$ for training and $400$ for testing.

For each initial condition $g_k(x)$, we first solve the equation without perturbations to obtain the clean trajectory. We then generate a noisy trajectory by introducing independent Gaussian perturbations of amplitude $\sigma\sqrt{\Delta t}$ at each fine time step, where $\sigma=0.35$. The clean trajectories are retained only for diagnostic evaluation and are not used during training or model selection.

\subsubsection{Federated Training}
\label{sec:burgers_auxdata}

The FedSON training procedure follows Algorithm~\ref{alg:fedson}. Each client first trains a local DeepONet using its $600$ noisy training trajectories. For this experiment, the training objective combines the mean squared error with a spectral penalty on high spatial wavenumbers,
\begin{equation}\label{eqn:burgers_loss}
\mathcal{L} = \mathcal{L}_{\mathrm{MSE}} + \lambda_{\mathrm{s}} \sum_{k \ge k_{\mathrm{lo}}} \bigl| \hat{u}_k^{\mathrm{pred}}-\hat{u}_k^{\mathrm{target}} \bigr|^2,
\end{equation}
where $\hat{u}_k$ denotes the $k$th spatial Fourier coefficient. We use $\lambda_{\mathrm{s}}=0.01$ and $k_{\mathrm{lo}}=5$ for all clients. The same spectral penalty ($\lambda_{\mathrm{s}} = 0.01$, $k_{\mathrm{lo}} = 5$) is used in all three deterministic settings: heterogeneous FedAvg, auxiliary FedAvg, and centralized training. In the heterogeneous and centralized runs, every training sample is real noisy data, and the penalty is applied to all samples; in the auxiliary run, it is applied only to the pseudolabel samples, whose smooth model-generated targets have negligible high-wavenumber content, and not to each client's own noisy samples, whose high-$k$ energy is genuine observational noise. The penalty is the mean of $|\hat{u}_k^{\mathrm{pred}} - \hat{u}_k^{\mathrm{target}}|^2$ over all retained wavenumbers $k \ge k_{\mathrm{lo}}$, all $N_t$ time steps, and all samples in the minibatch, so it is weighted uniformly across space, time, and batch.

The locally trained DeepONets are then used to construct the auxiliary datasets following Section~\ref{sec:fedson-aux}. Each client obtains an augmented dataset containing $3000$ training pairs, consisting of $600$ original noisy samples
and $2400$ pseudolabeled samples generated from the other four clients' initial conditions. The global DeepONet is initialized with new random parameters and trained using FedAvg on the augmented datasets.

\subsubsection{Deterministic Operator Results}
\label{sec:burgers_auxdata_results}
\begin{table}[h!]
\caption{[1D Burgers Eq.] Per-client test MSE for heterogeneous FedAvg, auxiliary FedAvg, and centralized DeepONet training. Centralized results are reported only against noisy
test trajectories.}
\label{tbl:burgers_mse}
\centering
\begin{tabular}{|c|c|c|c|c|c|c|} 
\hline
\multicolumn{2}{|c|}{} & \multicolumn{5}{|c|}{Client ID} \\
\hline
\multicolumn{2}{|c|}{Testing MSE ($10^{-5}$)} & 0 & 1 & 2 & 3 & 4 \\
\hline
\multirow{2}{*}{True Trajectory} & Hetero & 5982.7 & 27632 & 39939 & 40134 & 38546 \\
\cline{2-7}
& Auxi & 6.5674 & 77.332 & 147.13 & 478.05 & 716.43 \\
\hline
\multirow{3}{*}{Noisy Data} & Hetero & 6149.0 & 27803 & 40097 & 40304 & 38732 \\ 
\cline{2-7}
& Auxi & 178.50 & 248.47 & 325.89 & 653.06 & 892.22 \\
\cline{2-7}
& Central & 175.28 & 557.23 & 663.93 & 349.21 & 236.64 \\
\hline
\end{tabular}
\vspace{-0.2cm}
\end{table}

We first evaluate the deterministic DeepONet before proceeding to the stochastic component. Table~\ref{tbl:burgers_mse} compares the per-client test errors under heterogeneous FedAvg, auxiliary FedAvg, and centralized training. Consistent with the heat equation experiment, heterogeneous FedAvg produces substantially larger errors. The auxiliary data strategy significantly improves prediction accuracy, achieving errors of the same order of magnitude as centralized training. These results demonstrate that the auxiliary approach remains effective for the nonlinear Burgers' equation.
\begin{figure}[h!]
\centering
\includegraphics[width = 0.95\textwidth]{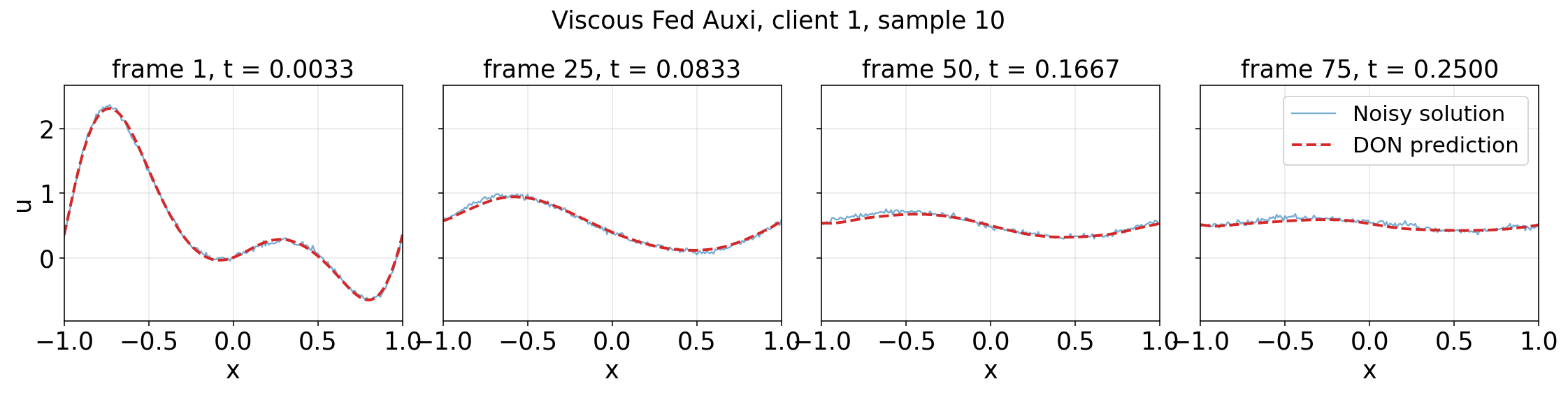}
\caption{[1D Burgers Eq.] Federated DeepONet with auxiliary data on the Burgers' equation: prediction on a representative test trajectory.}
\label{fig:burgers_fedavg_auxiliary-flat}
\end{figure}

Figure~\ref{fig:burgers_fedavg_auxiliary-flat} illustrates the prediction of the auxiliary-trained DeepONet on a representative test trajectory. The predicted solution closely
follows the clean trajectory throughout the time interval. The resulting DeepONet is then frozen and used as the deterministic component in the subsequent SON training.

\subsubsection{Stochastic Operator Network}
\label{sec:burgers_son}

We next train the SON using the federated DeepONet obtained in the previous subsection. The DeepONet parameters are frozen during SON training, which uses only the noisy trajectories, while the clean trajectories are reserved for diagnostic evaluation.
\begin{figure}[h!]
\centering
\includegraphics[width = 0.95\textwidth]{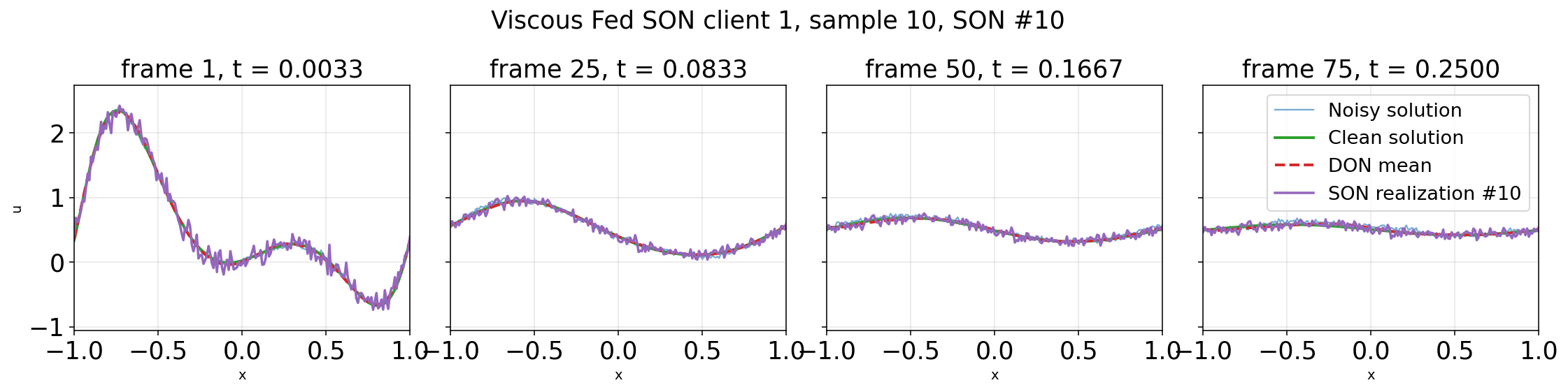}
\caption{[1D Burgers Eq.] Representative federated SON realization compared with the noisy trajectory, clean solution, and frozen DeepONet prediction.}
\label{fig:burgers_son_flat}
\vspace{-0.4cm}
\end{figure}

\begin{figure}[h!]
\centering
\begin{minipage}{0.49\textwidth}
\centering
\includegraphics[scale = 0.32]{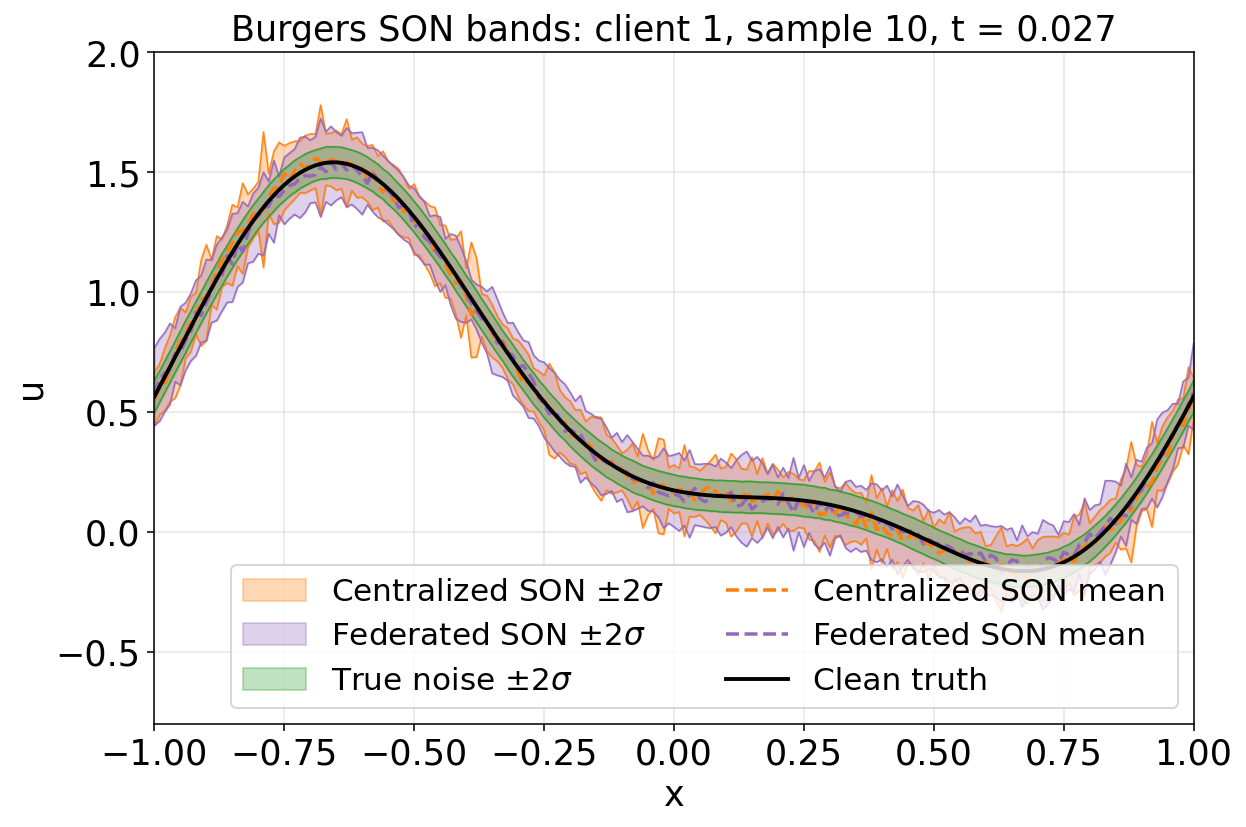}
\end{minipage}%
\begin{minipage}{0.49\textwidth}
\centering
\includegraphics[scale = 0.32]{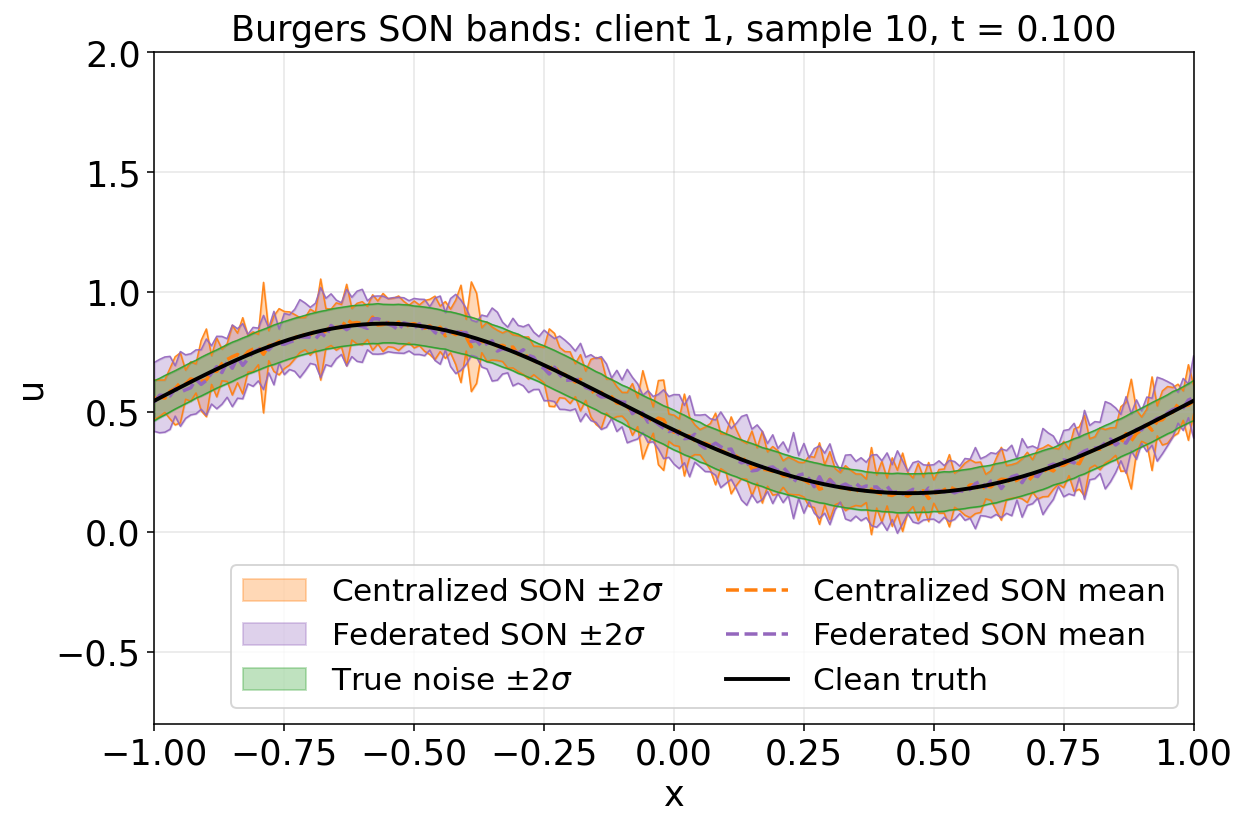}
\end{minipage}

\begin{minipage}{0.49\textwidth}
\centering
\includegraphics[scale = 0.32]{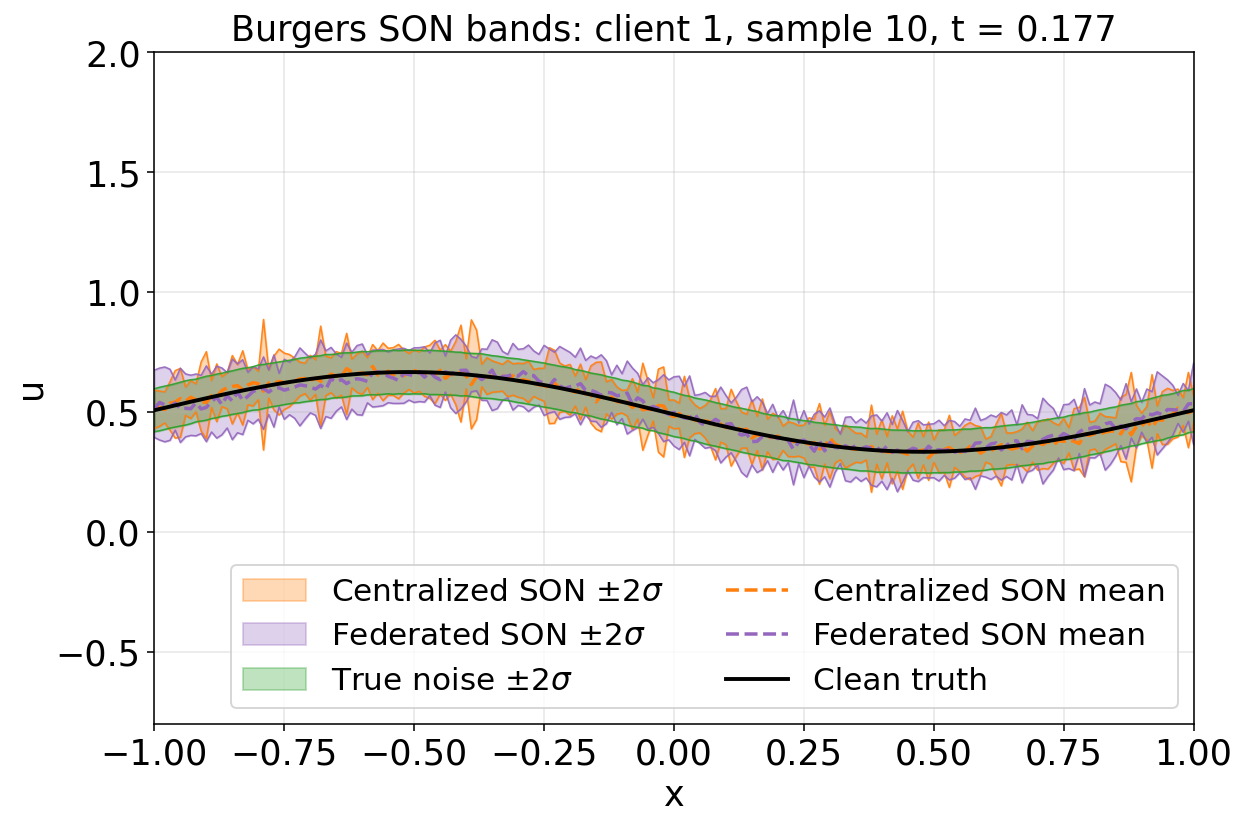}
\end{minipage}%
\begin{minipage}{0.49\textwidth}
\centering
\includegraphics[scale = 0.32]{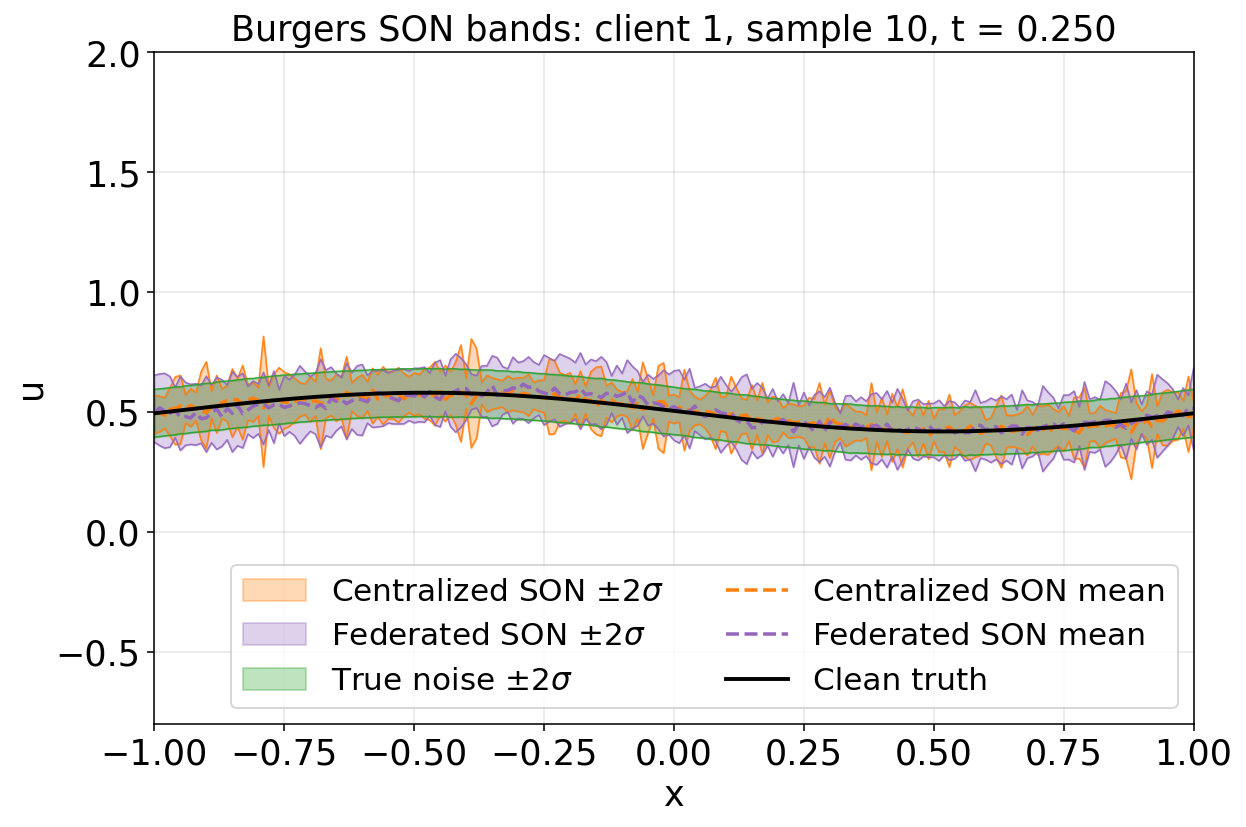}
\end{minipage}
\caption{[1D Burgers Eq.] Pointwise $\pm 2\sigma$ uncertainty bands for the
true data noise (green), federated SON (purple), and centralized SON (orange) at $t=0.027$, $0.10$, $0.177$, and $0.25$ for client~$1$, sample~$10$. The bands are centered on the clean solution or corresponding DeepONet predictions, with standard deviations estimated from $20$ SON realizations and pooled across clients and test samples at each spatial point.}
\label{fig:burgers_son_band}
\end{figure}

The SON is trained using the stochastic maximum principle introduced in Section~4. To calibrate the magnitude of the stochastic fluctuations, we initialize the noise scale using the standard deviation of the residuals between the noisy training trajectories and the frozen DeepONet predictions. At each Euler--Maruyama step, the diffusion output is normalized to unit root mean square before applying the calibrated noise scale, which remains fixed during training.  The diffusion network then learns the spatial and temporal structure of the stochastic correction. At inference, we generate $20$ independent SON realizations for each test initial condition to estimate the predictive statistics.

Figure~\ref{fig:burgers_son_flat} shows a representative SON realization for a test initial condition, together with the noisy trajectory, the true solution, and the frozen DeepONet mean. The SON realization fluctuates about the DeepONet mean and closely follows the noisy trajectory at every time slice. Because the noise amplitude is small relative to the magnitude of the solution, the individual realizations are difficult to distinguish from the true solution by eye; a quantitative comparison of the noise statistics is therefore required to assess the calibration of the learned uncertainty.

Across the time horizon, the federated SON produces fluctuations whose standard deviation is at the same level as the true data noise, confirming that the learned uncertainty is calibrated to the correct magnitude.



Figure~\ref{fig:burgers_son_band} compares the pointwise uncertainty bands of the federated SON, centralized SON, and true data noise at four times spanning the horizon. At every time, the federated and centralized SON bands closely track the true-noise band across the spatial domain, with the federated band only slightly wider than the centralized one. The calibration therefore holds as the solution evolves, not merely at a single time, indicating that the federated SON captures the spatial and temporal variation of the uncertainty despite the heterogeneous client distributions.

\subsection{Two-Dimensional Advection-Diffusion Equation}\label{sec:AD2D}


Finally, we investigate the performance of FedSON on a two-dimensional advection-diffusion equation. Building on the previous experiments, we evaluate the auxiliary data strategy under heterogeneous forcing distributions and examine the stochastic model's ability to characterize uncertainty in two-dimensional solution fields. The training data are generated using a mixed-hybrid finite element method (MHFEM), providing an additional test of the framework with a different numerical discretization.

\subsubsection{Governing Equation}\label{sec:ad2d-governing-equation}

Let $u(x,y,t)$ denote the scalar field on the unit square $\Omega = (0,1)^2$ over the time horizon $t \in [0,T]$ with $T = 1$. The field evolves according to the forced advection-diffusion equation
\begin{equation}
\label{eq:ad2d-pde}
\frac{\partial u}{\partial t} \;+\; \mathbf{b}\cdot\nabla u \;=\; \varepsilon\,\Delta u \;+\; 5\, g(t)\,f(x,y), \qquad (x,y) \in \Omega,\; t \in (0,T],
\end{equation}
where $\mathbf{b}(x,y)$ is a prescribed velocity field, $\varepsilon =0.05$ is the diffusivity, and $f(x,y)$ is the client dependent spatial forcing constructed in Section~\ref{sec:ad2d-forcing}. To close the system, we impose homogeneous Dirichlet boundary conditions and a zero initial condition,
\begin{equation}\label{eq:ad2d-bcic}
u = 0 \ \text{ on } \partial\Omega \times [0,T], \qquad u(x,y,0) = 0 \ \text{ on } \Omega,
\end{equation}
so that the entire solution is generated by the forcing: the trajectory that distinguishes one sample from another is driven purely by $f$ rather than by an initial state. The velocity field $\mathbf{b}$ is a fixed, piecewise-constant four-quadrant field shared across all clients and samples.

To avoid a rapid transition to a steady state under constant forcing, we introduce a time-dependent forcing factor
\begin{equation}\label{eq:ad2d-ramp}
g(t) = 1-e^{-t/\tau}, \qquad \tau=0.18,
\end{equation}
which allows the transient dynamics to develop more gradually over the time interval. This factor is fixed across all clients and samples and is not provided as an input to the models. As in the previous experiments, additive Gaussian noise is introduced at each time step, and the resulting noisy trajectories are used for training, while the deterministic solutions are retained only for diagnostic evaluation.



The training trajectories are generated by solving Eq.~\eqref{eq:ad2d-pde} using the mixed-hybrid finite element method (MHFEM) over a rectangular grid as described in~\cite{Brunner2014}, together with backward Euler time stepping. We use a uniform $40\times40$ spatial mesh and $N_t=80$ time steps of size $\Delta t=1/80$.

\subsubsection{Client Dependent Forcing}\label{sec:ad2d-forcing}

The federated setup comprises $C = 5$ clients. The forcing is expressed using tensor-product Chebyshev polynomials of the first kind, evaluated at the transformed coordinates $\hat{x} = 2x - 1$ and $\hat{y} = 2y - 1$. A forcing sample is a linear
combination
\begin{equation}\label{eq:ad2d-forcing}
    f(x,y) \;=\; \sum_{(i,j)} c_{ij}\, T_i(\hat{x})\, T_j(\hat{y}),
\end{equation}
over a two-dimensional index grid of modes, with the coefficient matrix $c = (c_{ij})$ supported only on the modes allotted to the client that generated the sample. We retain modes up to degree $5$ in each coordinate, i.e.\ a $6 \times 6$ coefficient grid $\{T_i(\hat{x})T_j(\hat{y})\}_{i,j=0}^{5}$.

To construct heterogeneous client datasets, we partition the two-dimensional Chebyshev modes into disjoint groups. Specifically, each client is assigned three modes according to
\begin{equation}\label{eq:ad2d-partition}
\mathcal{I}_k
= \big\{(k,k),\; (k,k+1),\; (k+1,k)\big\},
\qquad k=0,1,\dots,C-1.
\end{equation}
Each group consists of one diagonal mode and two adjacent off-diagonal modes, with the corresponding reflected modes assigned to the same client. The resulting index sets are pairwise disjoint, ensuring that no Chebyshev mode is shared between clients.


For a sample drawn by client $k$, the active coefficients $\{c_{ij} : (i,j) \in \mathcal{I}_k\}$ are drawn independently from the uniform distribution $\mathcal{U}(-1,1)$ with a fixed client specific seed, while all remaining coefficients are set to zero. Consistent with the gain in Eq.~\eqref{eq:ad2d-pde}, the forcing carries a fixed factor of $5$, the same amplification used in the heat experiment. Unlike the heat case, where the network is supplied with the unscaled field, here the branch network reads the scaled forcing $5f$ directly. For each client, we draw $N = 1000$ forcing samples, split them into $600$ training and $400$ testing, and propagate each through the stochastic MHFEM solver of Section~\ref{sec:ad2d-governing-equation} with noise level $\sigma = 0.35$ to produce the corresponding noisy trajectory. The resulting per client tensors are the forcing fields $f \in \mathbb{R}^{N \times N_y \times N_x}$ and the noisy trajectories $\tilde{u} \in \mathbb{R}^{N \times N_t \times N_y \times N_x}$, stored separately for training and testing.

\subsubsection{Model Architecture}\label{sec:ad2d-arch}



The deterministic operator is based on the time-marching DeepONet architecture introduced in Section~\ref{sec:heat-arch}, adapted to two-dimensional fields. Since the initial condition is identically zero and the solution is driven entirely by the forcing term, the branch network takes the spatial forcing $5 f(x,y)$ as its input. The trunk network receives the spatial coordinates $(x,y)$, while the decoder reconstructs the solution field from the latent representation.

To account for the continuous influence of the forcing throughout the time interval, we modify the latent propagator by supplying the branch encoding of the forcing term at every time step, together with the current latent state and the normalized step index. This forcing-conditioned propagator allows the latent dynamics to retain direct access to the forcing information during time integration. The same architecture is used for all training regimes to ensure a consistent comparison.

The FedSON training procedure follows Algorithm~\ref{alg:fedson}. Each client first trains a local DeepONet to generate pseudolabels for the auxiliary data construction described in Section~\ref{sec:fedson-aux}. The resulting augmented dataset for each of the $C=5$ clients contains $3000$ samples, consisting of $600$ original noisy samples and $2400$ pseudolabeled samples from the other four clients. Federated training is then performed using the augmented datasets.


\subsubsection{Deterministic Operator Results}\label{sec:ad2d-don-results}

We evaluate the deterministic DeepONet under four training settings: heterogeneous FedAvg on disjoint client distributions, homogeneous FedAvg on a stratified IID partition, the proposed auxiliary data strategy, and centralized training using the pooled data from all clients. All four use the same forcing-conditioned architecture and training settings. To assess the accuracy of the deterministic predictions, we report the relative $L_2$ error against the clean solution trajectories, which are retained only for diagnostic evaluation and are not used during training or model selection.

Table~\ref{tab:ad2d-don} reports the relative $L_2$ errors for each client and the pooled results. Heterogeneous FedAvg produces a pooled error of $0.729$, with particularly large errors for clients $3$ and $4$, indicating substantial performance degradation under the heterogeneous data partition. In contrast, the auxiliary data strategy reduces the pooled error to $0.041$, compared with $0.130$ for homogeneous FedAvg and $0.051$ for centralized training.

The auxiliary approach achieves the lowest pooled error among the four settings. One possible explanation for its improvement over homogeneous FedAvg is the difference in data allocation: each augmented dataset retains all $600$ original samples from the corresponding client and incorporates pseudolabels from the other clients, whereas the homogeneous partition distributes the available samples across the different forcing distributions. These results demonstrate that the auxiliary data strategy can substantially improve federated operator learning under heterogeneous data partitions without exchanging the original solution trajectories.

\begin{table}[h!]
\centering
\caption{[2D Adv-Diff Eq.] Per client test error for the four DeepONet settings on the two-dimensional advection--diffusion equation, measured as the relative $L_2$ error against the clean trajectories (used for diagnostics only). All four settings use the same forcing-conditioned DeepONet architecture. The homogeneous setting serves as a reference based on redistributed original data. Best deployable result per row in bold.}
\label{tab:ad2d-don}
\setlength{\tabcolsep}{6pt}
\begin{tabular}{@{}c c c c c@{}}
\hline
Client & Heterogeneous & Homogeneous (reference) & Auxiliary (ours) & Centralized\\
\hline
$0$ & $0.7257$ & $0.1114$ & $\mathbf{0.0235}$ & $0.0342$\\
$1$ & $0.5254$ & $0.1426$ & $\mathbf{0.0600}$ & $0.0382$\\
$2$ & $0.4504$ & $0.1897$ & $\mathbf{0.0635}$ & $0.0961$\\
$3$ & $1.0475$ & $0.1226$ & $\mathbf{0.0721}$ & $0.0433$\\
$4$ & $1.4577$ & $0.0800$ & $\mathbf{0.0544}$ & $0.0494$\\
\hline
pooled & $0.7288$ & $0.1296$ & $\mathbf{0.0405}$ & $0.0509$\\
\hline
\end{tabular}
\end{table}

\begin{figure}[h!]
\centering
\includegraphics[width=0.95\textwidth]{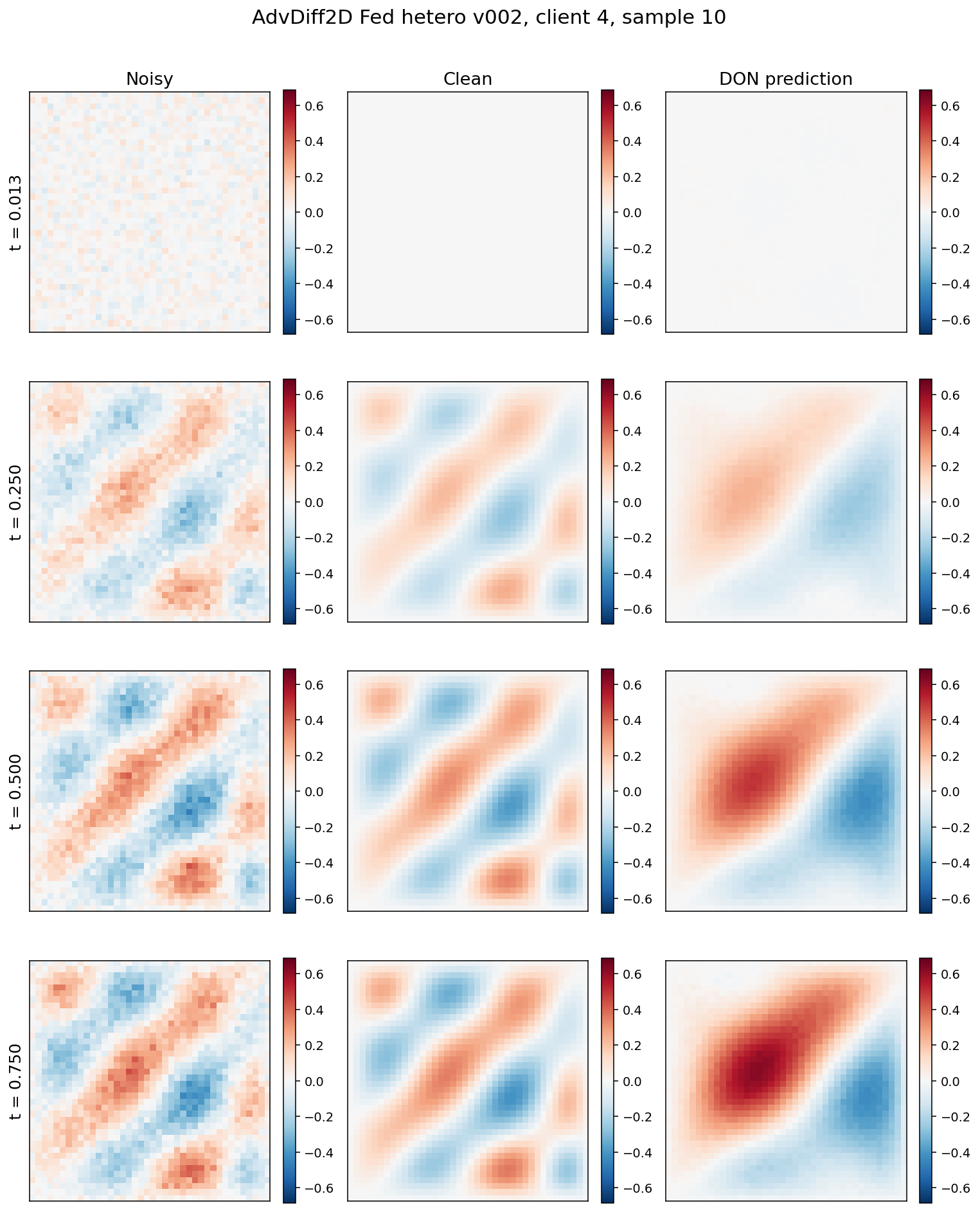}
\caption{[2D Adv-Diff Eq.] Heterogeneous FedAvg predictions for client~$4$, test sample~$10$. Rows show four time levels; columns show noisy data, clean solutions, and DeepONet predictions.}
\label{fig:ad2d-don-hetero}
\end{figure}

\begin{figure}[h!]
\centering
\includegraphics[width=0.95\textwidth]{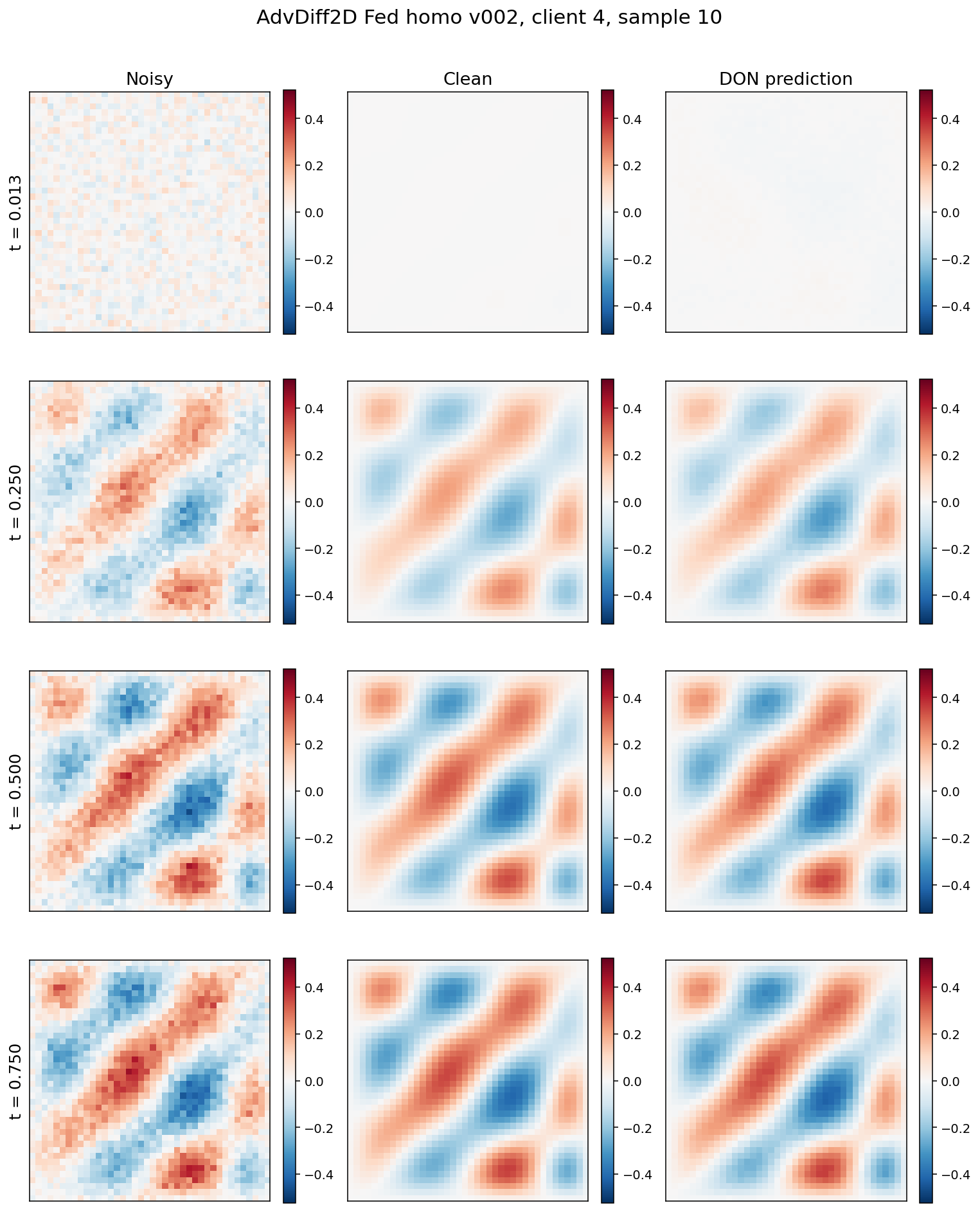}
\caption{[2D Adv-Diff Eq.] Homogeneous FedAvg predictions for client~$4$, test sample~$10$. The layout follows Figure~\ref{fig:ad2d-don-hetero}.}
\label{fig:ad2d-don-homo}
\end{figure}

\begin{figure}[h!]
\centering
\includegraphics[width=0.95\textwidth]{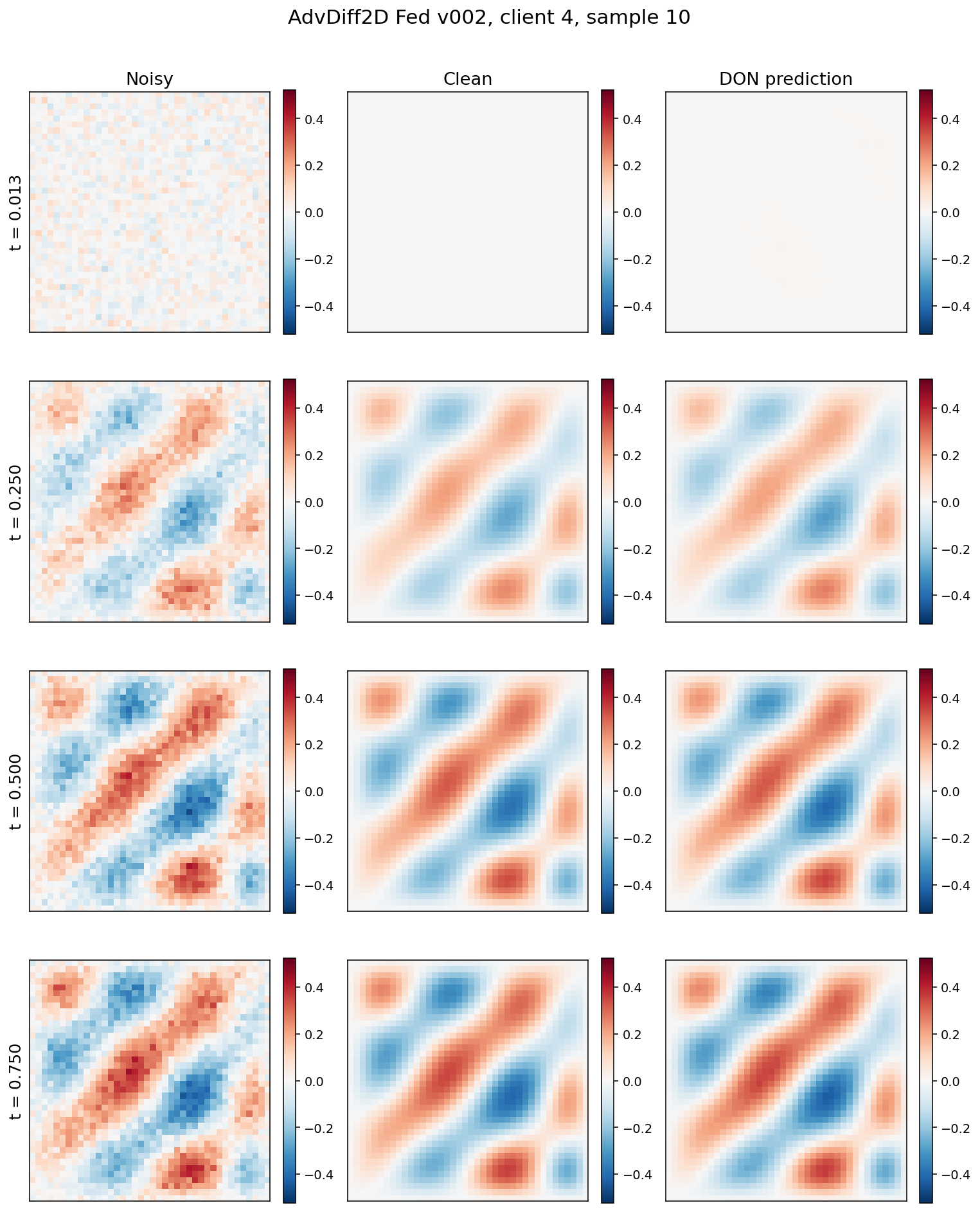}
\caption{[2D Adv-Diff Eq.] Auxiliary FedAvg predictions for client~$4$,
test sample~$10$. The layout follows Figure~\ref{fig:ad2d-don-hetero}.}
\label{fig:ad2d-don-aux}
\end{figure}

Finally, figures~\ref{fig:ad2d-don-hetero}-\ref{fig:ad2d-don-aux} show the predicted solution fields at four time levels for client~$4$, test sample~$10$, together with the noisy data and clean solutions. Under heterogeneous training (Fig.~\ref{fig:ad2d-don-hetero}), the prediction fails to reproduce the spatial structure of the solution, consistent with the large relative $L_2$ error reported in Table~\ref{tab:ad2d-don}. In contrast, both homogeneous FedAvg (Fig.~\ref{fig:ad2d-don-homo}) and the auxiliary approach (Fig.~\ref{fig:ad2d-don-aux}) recover the overall solution structure, with the auxiliary prediction showing closer agreement with the clean solution.


\subsubsection{Stochastic Operator Network}\label{sec:ad2d-son}


In Phase~II, the federated DeepONet obtained from Phase~I is frozen, and the SON is trained to characterize the stochastic fluctuations around its deterministic prediction. The stochastic dynamics are discretized using $N_\ell=20$ Euler-Maruyama steps with an Ornstein-Uhlenbeck drift toward the DeepONet prediction and a learnable low-rank diffusion operator consisting of $r=5$ networks. The diffusion is trained using the Hamiltonian-based formulation described in Section~\ref{sec:ol-son}. At inference, we generate $10$ SON realizations for each test input to estimate the predictive statistics.

The target noise level is estimated from the standard deviation of the residual between the noisy training trajectories and the frozen DeepONet prediction. To maintain this noise level during training, the diffusion increment is normalized to unit root mean square for each sample at every SDE step before applying the calibrated scaling factor. This prevents the magnitude of the diffusion network output from decreasing during training while allowing the network to learn the spatial and temporal structure of the stochastic fluctuations.

\begin{figure}[h!]
\centering
\includegraphics[width=0.95\textwidth]{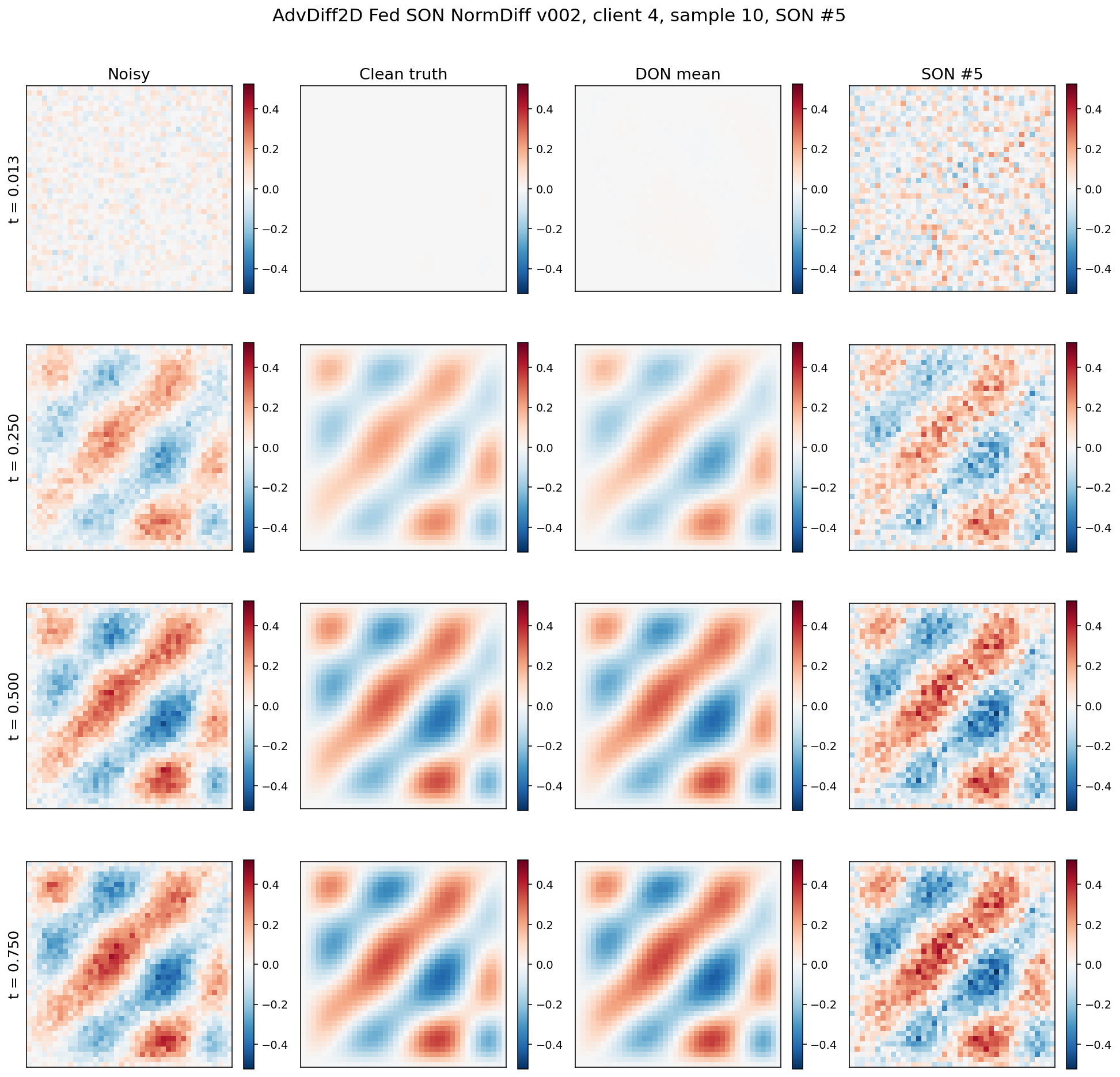}
\caption{[2D Adv-Diff Eq.] Federated SON prediction for client~$4$, sample~$10$, on the two-dimensional advection--diffusion equation. Rows are four time levels; columns are the noisy data, the clean trajectories, the frozen DeepONet mean, and one SON realization. The realization fluctuates about the DeepONet mean with a spread matching the noisy data, while the mean tracks the clean field.}
\label{fig:ad2d-son}
\end{figure}

Figure~\ref{fig:ad2d-son} compares a representative SON realization with the noisy trajectory, clean solution, and frozen DeepONet prediction at four time levels for client~$4$, sample~$10$. The DeepONet prediction captures the overall solution structure, while the SON introduces stochastic fluctuations around it. We next examine the corresponding uncertainty estimates through spatial cross-sections of the solution field.

\begin{figure}[h!]
\centering
\includegraphics[width=0.95\textwidth]{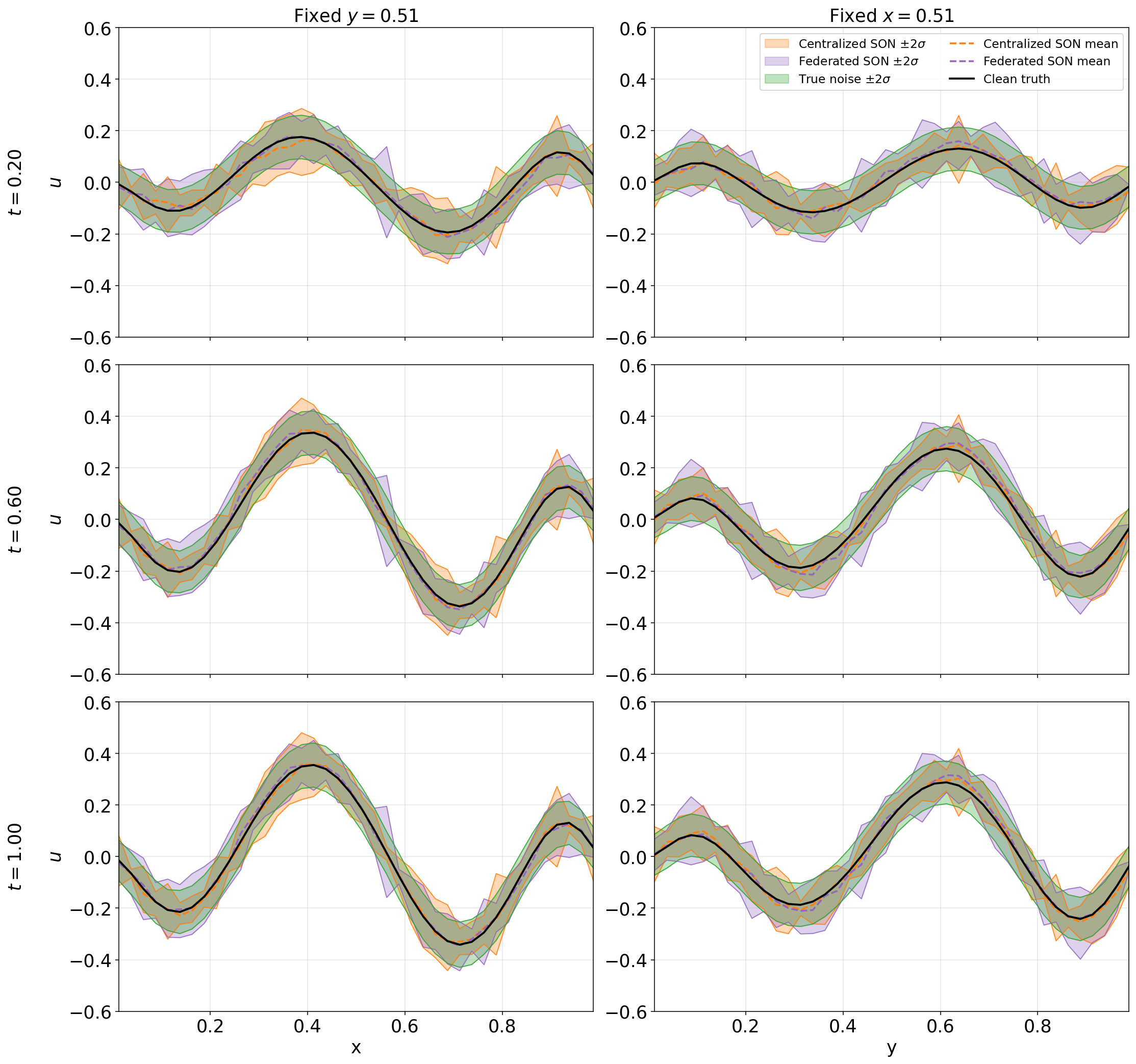}
\caption{[2D Adv-Diff Eq.] Pointwise uncertainty bands for client $4$, sample $10$, on two orthogonal cross-sections of the two-dimensional field at three time levels ($t = 0.2$, $0.6$, $1.0$ from top to bottom). The left column is a horizontal cut at fixed $y$ (plotted against $x$) and the right column a vertical cut at fixed $x$ (plotted against $y$). The black curve is the clean solution and the dashed curves are the empirical SON means; the shaded regions are the $\pm 2\sigma$ envelopes of the true data noise, the federated SON, and the centralized SON. Each band width is the population standard deviation pooled across clients and test samples over the full field at that time, sliced on the plotted cross-section.}
\label{fig:ad2d-son-band}
\end{figure}

Figure~\ref{fig:ad2d-son-band} compares the uncertainty bands of the federated SON, centralized SON, and true data noise along two orthogonal cross-sections at three time levels. The federated and centralized SON models produce bands of comparable width, with both showing reasonable agreement with the true-noise bands across the selected spatial locations and times. Since Phase~II uses only local noisy trajectories and aggregates model parameters through FedAvg, these results suggest that the federated SON can reproduce important spatial and temporal features of the stochastic fluctuations without exchanging the original solution trajectories.


\section{Conclusion}\label{sec:conclusion}

We have introduced FedSON, a federated framework for stochastic operator learning that recovers both the mean solution operator and a calibrated estimate of the predictive uncertainty under severe data heterogeneity, without any client ever sharing its solution trajectories. Built on the two-phase stochastic operator network, FedSON trains a deterministic DeepONet to capture the mean and a stochastic diffusion network to capture the fluctuations about it, each under federated averaging, and augments the clients' disjoint input supports through a pseudolabel exchange that communicates models and input functions but no solution data. Across the heat, viscous Burgers, and one and two dimensional advection diffusion equations, naive federated averaging fails on the fully disjoint partition, whereas the auxiliary pipeline matches or exceeds the homogeneous reference and comes within a small factor of a centralized baseline trained on the pooled data, all while keeping every client's solution data local. In the second phase, the federated stochastic operator network reproduces the measured noise level pointwise across space and time and attains an uncertainty estimate comparable to that of a centralized model with unrestricted access to every client's data.

Several directions remain open. The present framework adopts standard federated averaging as the aggregation rule and does not incorporate complementary privacy mechanisms such as secure aggregation or differential privacy; combining pseudolabel exchange with formal privacy guarantees is a natural next step. The pseudolabel construction also communicates each client's input functions in full, and quantifying the residual information this reveals, as well as extending the approach to settings where the input functions themselves are sensitive, would broaden its applicability. Finally, the experiments here use a fixed client count and a controlled, disjoint partition; evaluating FedSON on larger client populations, on real distributed scientific datasets, and on higher-dimensional or more complex operators are important avenues for future work.

\bibliographystyle{elsarticle-harv} 
\bibliography{reference}
\end{document}